\documentclass{article}

\PassOptionsToPackage{numbers}{natbib}
\usepackage[final]{neurips_2023}
\usepackage[utf8]{inputenc} % allow utf-8 input
\usepackage[T1]{fontenc}    % use 8-bit T1 fonts
\usepackage[bookmarks=false,colorlinks]{hyperref}
\usepackage{url}            % simple URL typesetting
\usepackage{booktabs}       % professional-quality tables
\usepackage{amsfonts}       % blackboard math symbols
\usepackage{nicefrac}       % compact symbols for 1/2, etc.
\usepackage{microtype}      % microtypography
\usepackage{xcolor}         % colors

\usepackage{graphicx}
\usepackage{amsmath}
\usepackage{algorithmic}
\usepackage{float}

\title{Cooperative Multi-Objective Reinforcement Learning for Traffic Signal Control and Carbon Emission Reduction }

\author{%
Cheng Ruei Tang\textsuperscript{\rm 1},Jun Wei Hsieh\textsuperscript{\rm 2},Shin You Teng\textsuperscript{\rm 3},\thanks{Use footnote for providing further information
    about author (webpage, alternative address)---\emph{not} for acknowledging
    funding agencies.} \\
  \textsuperscript{\rm 123}National Yang Ming Chiao Tung University,\\
  \textsuperscript{\rm 13}Institute of Computational Intelligence student,\textsuperscript{\rm 2}Professor of Computational Intelligence\\
  \texttt{\textsuperscript{\rm 1}ray86224@gmail.com,\textsuperscript{\rm 2}ai@nycu.edu.tw,jwhsieh@nycu.edu.tw}, \texttt{\textsuperscript{\rm 3}tengyoyo.ai10@nycu.edu.tw}
 \\
}

\begin{document}

\maketitle

\begin{abstract}
\vspace{-0.3cm}
% We propose an actor-critic reinforcement learning (RL) solution for adaptive traffic signal control optimization. The existing traffic signal control systems still rely on oversimplified rule-based methods. Even the latest RL-based traffic control methods are far from optimal and unstable in general cases. Specifically, the periodicity of green/red light alternations can be considered as a prior for better planning of each agent in policy optimization and peer forecasting.
% To better learn such adaptive and predictive priors, we propose a cooperative multi-objective architecture with age-decaying weights to better estimate multiple reward terms for traffic signal control optimization. Our model is termed Multi-Objective Multi-Agent Deep Deterministic Policy Gradient (MOMA-DDPG). Two types of agents learn to maximize rewards of different goals: one for local traffic optimization at each intersection and the other for global traffic throughput optimization.  The evaluation is performed using real-world traffic data collected using traffic cameras from an Asian country. Although a global agent is included, our method is a decentralized solution since this agent is not longer needed during the inference stage. The results demonstrate its efficacy compared to those of SoTA methods in all performance metrics. The proposed MOMA-DDPG system can reduce not only the waiting time but also carbon emission.
%after gpt
Existing traffic signal control systems rely on oversimplified rule-based methods, and even RL-based methods are often suboptimal and unstable. To address this, we propose a cooperative multi-objective architecture called Multi-Objective Multi-Agent Deep Deterministic Policy Gradient (MOMA-DDPG), which estimates multiple reward terms for traffic signal control optimization using age-decaying weights. Our approach involves two types of agents: one focuses on optimizing local traffic at each intersection, while the other aims to optimize global traffic throughput. We evaluate our method using real-world traffic data collected from an Asian country's traffic cameras. Despite the inclusion of a global agent, our solution remains decentralized as this agent is no longer necessary during the inference stage. Our results demonstrate the effectiveness of MOMA-DDPG, outperforming state-of-the-art methods across all performance metrics. Additionally, our proposed system minimizes both waiting time and carbon emissions. Notably, this paper is the first to link carbon emissions and global agents in traffic signal control.
\end{abstract}
\vspace{-0.8cm}

% \subsection{Figures}

% \begin{figure}
%   \centering
%   \fbox{\rule[-.5cm]{0cm}{4cm} \rule[-.5cm]{4cm}{0cm}}
%   \caption{Sample figure caption.}
% \end{figure}

\section{Introduction}
\vspace{-0.3cm}
Traffic signal control is a challenging real-world problem whose goal tries to minimize the overall vehicle travel time by coordinating the traffic movements at road intersections. Existing traffic signal control systems in use still rely heavily on manually designed rules which cannot adapt to dynamic traffic changes and cannot deal well with today's increasingly large transportation networks. Recent advance in reinforcement learning (RL), especially deep RL~\cite{alemzadeh2020adaptive,zheng2019diagnosing}, offers excellent capability to work with high dimensional data, where agents can learn a state abstraction and policy approximation directly from input states.  This paper explores the possibility of RL to on-policy traffic signal control with fewer assumptions.

In literature, there have been different RL-based frameworks~\cite{wei2021recent} proposed for traffic light control.  Most of them~\cite{zheng2019diagnosing,mannion2016experimental,pham2013learning,van2016coordinated,wei2018intellilight,arel2010reinforcement,calvo2018heterogeneous} are value-based and can achieve convergence in relatively easier steps. However, their vita problem is:  only discrete actions, states, and time spaces can be applied to.\footnote{
A recent method~\cite{ValueIter:Continuous:ICML2021} can handle value-iterations in continuous actions, states, and time but it has not been applied to traffic optimization yet.} For traffic optimization, this means the choice of next traffic light phase is constrained and limited from pre-defined discrete cyclic sequences of red/green lights. The solution of pre-defining time slots for the agent to simply determine actions is effective for optimization and simple for traffic control.  However, the time slots for each action to be executed are fixed and cannot reflect the real requirements to optimize traffic conditions.  Moreover, a small change in the value function will cause great effects on the policy decision.  To make decisions on continuous space, recent policy-based RL methods~\cite{chu2019multi,nishi2018traffic,mousavi2017traffic} become more popularly adopted in traffic signal control so that a non-discrete length of phase duration can be inferred.  However, its gradient estimation is strongly dependent on sampling and not stable, and thus easily trapped to a non-optimal solution. 
 
 To bridge the gaps between value-based and policy-based RL approaches, the actor-critic framework is widely adopted to stabilize the RL training process, where the policy structure is known as the actor and the estimated value function is known as the critic. The Deep Deterministic Policy Gradient method (DDPG)~\cite{lillicrap2015continuous} learns a $Q$-function and a policy concurrently, by using off-policy data and the Bellman equal to learn the $Q$-function, and then using the Q-function to learn the policy. DDPG retains the advantages of both the value-based and policy-based method, and can directly learn a deterministic policy mapping states to actions directly. Thus, there are some actor-critic frameworks proposed for traffic signal control.  For example,~\cite{pang2019deep,wu2020control}, DDPG was adopted to learn a deterministic policy mapping states to actions. However, it is a ``local-agent'' solution which is not optimized by trading off different local agents' requirements. More precisely, current RL-based solutions~\cite{pang2019deep,wu2020control,chu2019multi,nishi2018traffic,mousavi2017traffic} are ``locally'' derived from single agent or multi-agents. Each agent individually decides its policies and actions according to its rewards which often produce conflicts to other agents and make traffic congestion more serious in other intersections. The above RL-based frameworks lack of a global agent to cooperate all local agents and trade off their different requirements.   
 Another issue of traffic signal control is making 
 decisions not only on which action to be performed but also how long it should be performed. There are only few frameworks~\cite{aslani2017adaptive,aslani2018traffic} pre-defining several fixed time slots from which the agent can choose to determine the action period.  However, this pre-defining solutions are less flexible than an on-demand solution to better relieve traffic congestion.
 
 This paper develops a COoperative  Multi-objective Multi-Agent DDPG (COMMA-DDPG) framework for optimal traffic signal control. Current RL-based multi-agent methods use only local agents to search solutions and often produce conflicts to other agents.   The novelty of this paper is to introduce a global agent to cooperate with all local agents by trading off their requirements to increase the entire throughput.  To the best of our knowledge, this concept has not been explored in the literature for traffic signal control. Fig. 3 (in the supplement) shows an overview of our proposed COMMA-DDPG framework.  Each local agent focuses on learning the local policy using the intersection clearance as reward.  The global agent then optimizes the overall rewards measured by the total traffic waiting time. With the actor-critic framework, the global agent can optimize various information exchanges from local intersections so as to optimize the final reward globally.   Our COMMA-DDPG can select the best policy to control the periodic phases of traffic signals that maximize throughput.  The global agent is used only during the training stage and is no longer needed in the inference stage.  It may be problematic when training agents over a larger road network (more than 10 intersections), since both local and global agents take the information of all intersections as input.  To address this problem, for each local agent, 
 we adjust the global agent by sending information only from its nearby intersections as a training basis because intersections too far away are not so important. 
 Additionally, unlike other policy-based methods that can only return a fixed length from a predefined action pool, the COMMA-DDPG mechanism allows for determining not only the best policy but also a dynamic length for the next traffic light phase. Theoretical support for the convergence of our COMMA-DDPG approach is provided. Moreover, this paper is the first to establish a link between carbon emissions and global agents in traffic signal control.   Experimental results show that the COMMA-DDPG framework significantly improves overall waiting time, as well as reduces $CO_2$ emissions and carbon emissions.  Main contributions of this paper are summarized in the following~:  
\begin{itemize} %\itemsep -.5em
\item We propose a COMMA-DDPG framework that can effectively improve the traffic congestion problem, reduce travel time, and thus increase the entire throughput of the roads. 
 \vspace{-0.1cm}  
\item The global-agent design  can trade off different conflicts among local agent and give guides to each local agent so that better policies can be determined.       
\item The COMMA-DDPG method can determine not only the best policy, but also a dynamic length of the next traffic light phase.
\item The COMMA-DDPG method can reduce not only the waiting time, but also $CO_2$ and carbon emissions.
\item Extensive experiments on real-world traffic data and an open bechmark\cite{ault2021reinforcement} show that the COMMA-DDPG method  achieves SoTA results for effective and efficient traffic signal control. 
    
\end{itemize}
\begin{figure}
    
    \includegraphics[scale=0.25]{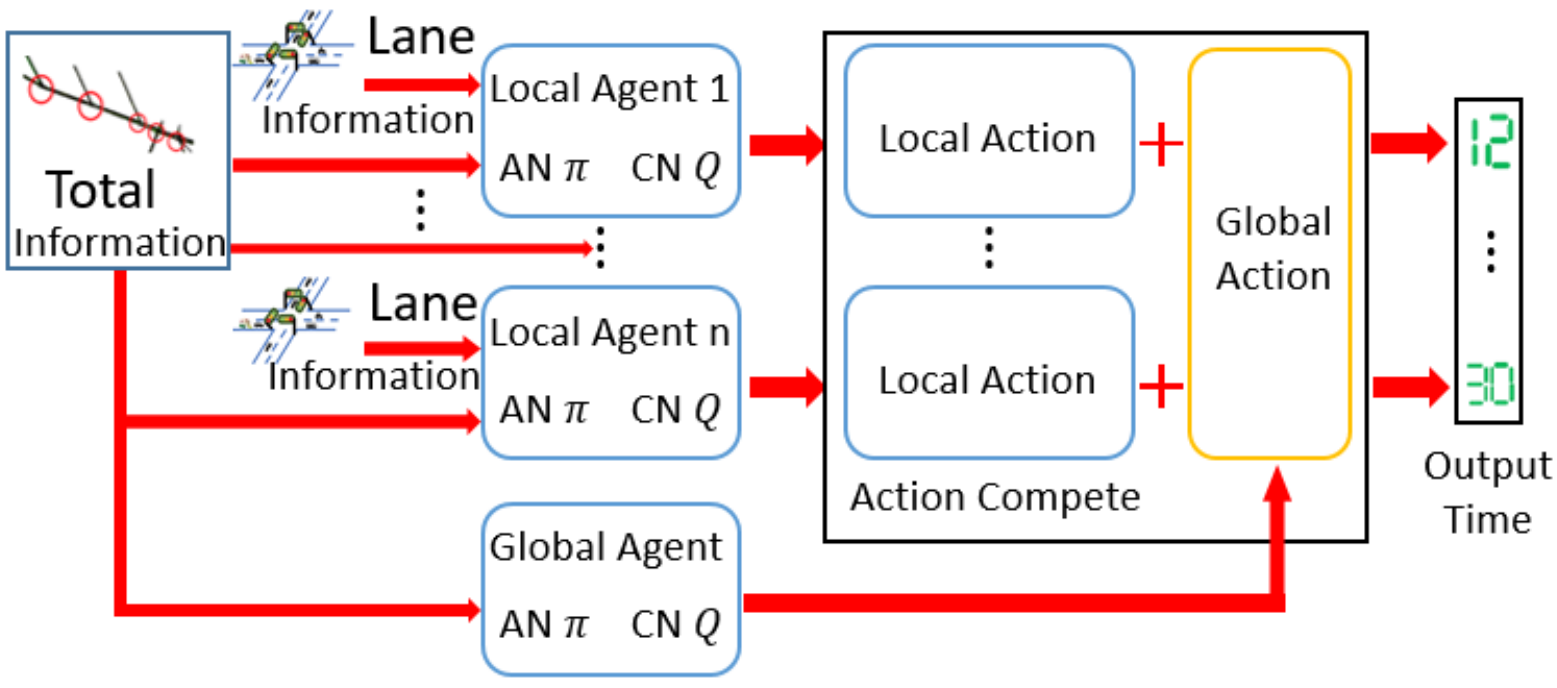}
    \includegraphics[scale=0.25]{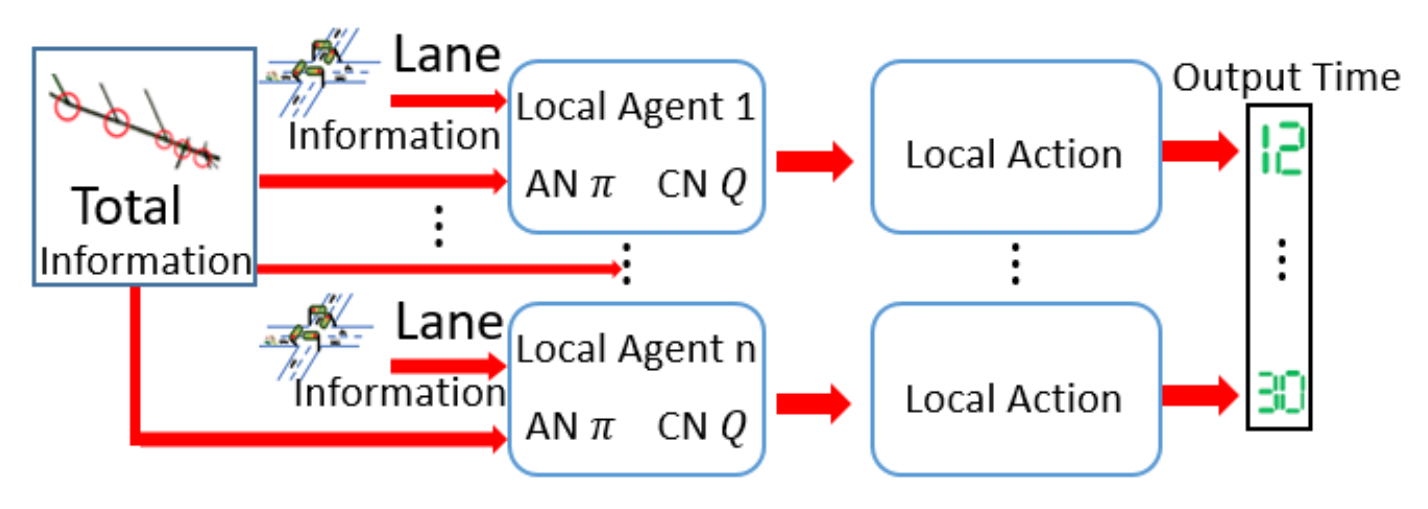}
    \vspace{-0.2cm}  
    
    \hspace{3 cm}
    {\footnotesize (a)} \hspace{6 cm}
    {\footnotesize (b)}
    % \vspace{-0.2cm}
    \caption{Architectures of COMMA-DDPG.
    (a) Architecture used in training stage.
    (b) Architecture used in inference stage.} 
    \vspace{-0.65cm}
    \label{fig:proces}
\end{figure}
%%%%%%%%%%%%%%%%%%%%%%%%%%%%%%%%%%%%%%%
 \vspace{-0.3cm}
\section{Related Work}
\label{sec:related}
\vspace{-0.4cm}

{\bf Traditional traffic control} methods can be categorized into three classes~: (1) fixed-time control~\cite{roess2004traffic}, (2) actuated control~\cite{fellendorf1994vissim,mirchandani2001real}, and (3) adaptive control~\cite{zheng2019learning,zheng2019diagnosing}. They are mainly based on human knowledge to design appropriate cycle length and strategies for better traffic control. The manual tasks involved will make parameter settings very cumbersome and difficult to satisfy different scenarios' requirements, including peak hours, normal hours, and off-peak hours. Fixed-time control is simply, easy, and thus becomes the most commonly adopted method in traffic signal control. Actuated control determines traffic conditions using predetermined thresholds; that is, if the traffic condition (e.g., the car queue length) exceeds a threshold, a green light will be issued accordingly. Adaptive control methods including~\cite{lowrie1990scats} determine the best signal phase according to current traffic conditions and thus can achieve more effective traffic optimization.

\vspace{-0.2cm}
{\bf RL traffic control:} Recent advances in RL shed light on the improvement of automatic traffic control.  There are two main approaches to solving traffic sign control problems, {\em i.e.}, value-based and policy-based. There is also a hybrid actor-critic (AC) approach~\cite{aslani2017adaptive}, which employs both value-based and policy-based searches. The value-based method first estimates the value (expected return) of being in a given state and then finds the best policy from the estimated value function.  One of the most widely used value-based methods is $Q$ learning~\cite{watkins1992q}.  The first $Q$-learning method applied to control traffic signals at street intersection is traced from~\cite{abdoos2011traffic}. However, in $Q$ learning, a large table should be created and updated to store the $Q$ values of each action in each state. Thus, it is both memory- and time-consuming and improper for problems with complicated states and actions.  Thus, various RL-based methods~\cite{zheng2019diagnosing,wei2019presslight,wei2019colight,aslani2017adaptive} have been proposed for traffic signal control.  Among them, Deep Q learning Network (DQN)~\cite{guo2014deep} is often adopted to estimate the $Q$ function.  However, the max operator in DQN uses the same values both to select and to evaluate an action.  The selection often makes values overestimated. The double DQN~\cite{van2016deep} decouples the selection from the evaluation by using two networks to solve this overestimation problem. The flaw in the value-based methods is that their quick convergences require a discrete action state.  The policy-based methods can directly optimize the desired policy with policy gradient for fast convergence. But there is a disadvantage that the policy-based method is a round update during the training process, so the training process would be long. The work of~\cite{casas2017deep} has verified the superior fitting power of the deep deterministic policy gradient under simplified traffic environment. AC approaches are the trend for traffic control.

RL can also be classified according to the adopted action schemes, such as: (i) setting the length of green light, (ii) choosing whether to change phase, and (iii) choosing the next phase. The DQN and AC methods are suitable for action schemas 2 and 3, but are not suitable for setting the length of the green line since the action space of DQN is discrete, and a lot of calculations will be wasted for the AC method. DDPG~\cite{lillicrap2015continuous} can solve continuous action spaces, which are more suitable for modeling the length of green light.  Thus, there are several DDPG-based RL frameworks~\cite{pang2019deep,wu2020control} proposed for traffic signal control. However, they focus on only a single intersection.  In practical environments, multi-agent deep deterministic policy gradient algorithms ~\cite{gupta2017cooperative,Lowe2017MMDDPG} will be another good choice fortraffic control to incorporate information from different agents for large-scale traffic scenarios.  For example, the CityFlow algorithm~\cite{Zhang2019CityFlow} is proposed  to control traffic signals in a city. However, the above multi-agent systems cannot provide a dynamic length of the next traffic light phase to inform the drivers.
%after gpt
% RL can be classified based on the adopted action schemes, which include setting the length of the green light, choosing whether to change the phase, and selecting the next phase. The DQN and AC methods are suitable for action schemes 2 and 3 but not for setting the length of the green light, as the action space of DQN is discrete and the AC method requires excessive calculations. DDPG \cite{lillicrap2015continuous}, on the other hand, can handle continuous action spaces, making it more suitable for modeling the length of the green light. As a result, several DDPG-based RL frameworks \cite{pang2019deep, wu2020control} have been proposed for traffic signal control. However, these frameworks mainly focus on a single intersection. In practical environments with large-scale traffic scenarios, multi-agent deep deterministic policy gradient algorithms \cite{gupta2017cooperative, Lowe2017MMDDPG} offer a better choice for traffic control as they can incorporate information from different agents. For instance, the CityFlow algorithm \cite{Zhang2019CityFlow} is designed to control traffic signals in a city. However, the aforementioned multi-agent systems cannot provide a dynamic length for the next traffic light phase to inform the drivers.
\vspace{-0.5cm}

%%%%%%%%%%%%%%%%%%%%%%%%%%%%%%%%%%%%%%%%%%%%%%%%%
\section{Background and Notations}
\label{sec:notations}
\vspace{-0.3cm}
The basic elements of a RL problem for traffic signal control can be formulated as a Markov Decision Process (MDP) mathematical framework of $< S,A,T,R,\gamma >$, with the following definitions:
\vspace{-0.2cm}
% \begin{itemize}[leftmargin=16pt]
\begin{itemize}
\setlength\itemsep{-0.028cm}
\item $S$ denotes the set of states, which is the set of all lanes containing all possible vehicles. $s_t  \in S$ is a state at time step $t$ for an agent.
\item $A$ denotes the set of possible actions, which is the duration of green light. In our scenarios, both duration lengths for a traffic cycle and a yellow light are fixed.  Then, once the state of green light is chosen, the duration of a red light can be determined.  At time step $t$, the agent can take an action $a_t$ from  $A$.
\item $T$ denotes the transition function, which stores the probability of an agent transiting from state $s_t$  to $s_{t+1}$  if the action $a_t$ is taken; that is, $T(s_{t+1}|s_{t},a_{t} ):S \times A \rightarrow S.$
\item 	$R$ denotes the reward, where at time step $t$, the agent obtains a reward $r_t$ specified by a reward function $R(s_{t},a_{t} )$ if the action $a_t$ is taken under state $s_t$. 
\item $\gamma$ denotes the discount, which controls the importance of the immediate reward versus future rewards, and also ensures the convergence of the value function, where $\gamma \in [0,1)$.
\end{itemize}
\vspace{-0.2cm}
At time-step $t$, the agent determines its next action $a_t$ based on the current state $s_t$.  After executing $a_t$, it will be transited to next state $s_{t+1}$ and receive a reward $r_{t} (s,a)$; that is, $r_{t} (s,a)= \mathbb{E}[R_{t}|s_{t}=s,a_{t}=a]$,  where $R_t$ is named as the one-step reward. The way that the RL agent chooses an action is named policy and denoted by $\pi$. Policy is a function $\pi (s)$ that chooses an action from the current state $s$; that is, $\pi (s):S \rightarrow A$. Our goal is to find such a policy to maximize the future reward 
% $G_t=\sum_{k=0}^{\infty} \gamma^k R_{t+k}$.\\
$G_t$:
$\vspace{-0.1cm}$
\begin{equation}
\label{equ:Gt}
    G_{t}=\Sigma_{k=0}^{\infty}\gamma^{k}R_{t+k}.
\vspace{-0.1cm}
\end{equation}
A value function $V(s_t)$ indicates how good the agent is in state $s_t$, $i.e.$, the expected total return of the agent starting from $s_t$. If $V(s_t)$ is conditioned on a given strategy $\pi$, it will be expressed by $V^{\pi}(s_{t})$; that is, $V^{\pi}(s_{t})=\mathbb{E}[G_{t}|s_{t}=s], \forall s_{t} \in S$.  The optimal policy $\pi^{*}$ at state $s_t$ can be found by  
% $\pi^* (s_{t})= \mathop{\arg\max}\limits_{\pi} V^{\pi} (s_t)$
$\vspace{-0.1cm}$
\begin{equation}
    \pi^{*}(s_t)=arg \max\limits_{\pi}V^{\pi}(s_t), 
\label{eq2}
\vspace{-0.1cm}
\end{equation}  
where $V^{\pi}(s_t)$ is the state-value function for a policy $\pi$.  Similarly, we can define the expected return of taking action $a$ in state $s_t$ under a policy $\pi$ denoted by a $Q$ function: 
% $Q^{\pi} (s_t ,a_t)=\mathbb{E}[G_t\mid s_t = s,a_t =a]$.
\begin{equation}
    Q^{\pi}(s_{t},a_{t})=\mathbb{E}[G_{t}|s_{t}=s,a_{t}=a].
\end{equation}
The relationship between $Q^{\pi}(s_t, a_t)$ and $V^{\pi} (s_t)$ is 
% ${{\rm{V}}^\pi }{\rm{(}}s) = \sum\limits_{a \in A} {\pi (a|s){Q^\pi }(s,a)}.$
\begin{equation}
{{\rm{V}}^\pi }{\rm{(}}s) = \sum\limits_{a \in A} {\pi (a|s){Q^\pi }(s,a)}. 
\vspace{-0.1cm}
\end{equation}
Then, the optimal $Q^{*}(s_{t},a)$ is  iteratively solved by 
% $Q^{*}(s_{t},a)=\max\limits_{\pi}Q^{\pi}(s_{t},a)$. 
\begin{equation}
Q^{*}(s_{t},a)=\max\limits_{\pi}Q^{\pi}(s_{t},a).
\vspace{-0.1cm}
\end{equation} 
With $Q$, the optimal policy $\pi^{*}$ in state $s_t$ can be found by:
% $\pi^{*}(s_t)=arg \max\limits_{a}Q^{*}(s_t,a) $. 
\begin{equation}
\pi^{*}(s_t)=arg \max\limits_{a}Q^{*}(s_t,a). 
\vspace{-0.1cm}
\end{equation} 
$Q^{*}(s_{t},a)$ is the sum of two terms: (i) the instant reward after a period of execution in the state $s_t$ and (ii) the discount expected future reward after the transition to the next state $s_{t+1}$. Then, we can use the Bellman equation\cite{bellmanequation} to express $Q^{*}(s_t,a)$ as follows: 
$\vspace{-0.1cm}$
% $Q^{*}(s_{t},a)=R(s_{t},a)+\gamma\mathbb{E}_{s_{t+1}}[V^{*}(s_{t+1})]$
\begin{equation}
Q^{*}(s_{t},a)=R(s_{t},a)+\gamma\mathbb{E}_{s_{t+1}}[V^{*}(s_{t+1})]. 
\label{eq:BellmanQ}
\vspace{-0.1cm}
\end{equation}  
$V^*(s_t)$ is the maximum expected total reward from state $s_t$ to the end.  It will be the maximum value of $Q^*(s,a)$ among all possible actions. Then, $V^*$ can be obtained from $Q^*$ as:
$\vspace{-0.1cm}$
% ${V^*}\left( {{s_t}} \right) = \mathop {\max }\limits_a {Q^*}\left( {{s_t},a} \right),\forall {s_t} \in S$.\\
\begin{equation}
{V^*}\left( {{s_t}} \right) = \mathop {\max }\limits_a {Q^*}\left( {{s_t},a} \right),\forall {s_t} \in S.
\vspace{-0.1cm}
\end{equation} 
{\bf Deep Q-Network (DQN):} In~\cite{hester2018deep,van2016deep}, a deep neural network is used to approximate the $Q$ function, which enables the RL algorithm to learn $Q$ well in high-dimensional spaces. Let $Q_{tar}$ be the targeted true value which is expressed as $Q_{tar}= r+\gamma \max\limits_{a'}Q(s',a';\theta)$.  In addition, let $Q(s,a;\theta)$ be the estimated value, where $\theta$ is the set of its parameters.   We define the loss function for training the DQN as:
$\vspace{-0.1cm}$
% $L(\theta)=\mathbb{E}_{s,a,r,s'}[(Q_{tar}-Q(s,a;\theta))^{2}]$.\\
\begin{equation}
L(\theta)=\mathbb{E}_{s,a,r,s'}[(Q_{tar}-Q(s,a;\theta))^{2}].
% \vspace{-0.1cm}
\end{equation}
In ~\cite{mnih2015human}, $Q_{tar}$ is often overestimated during training and results in the problem of unstable convergence of the $Q$ function.  
In~\cite{van2016deep}, a Double DQN (DDQN) was proposed to deal with this unstable problem by separating the DDQN into two value functions, so that there are two sets of weights $\theta$ and $\phi$ for parameterizing the original value function and the second target network, respectively.  The second DQN $Q_{tar}$ with parameters $\phi$ is a lagged copy of the first DQN $Q(s,q;\theta)$ that can fairly evaluate the $Q$ value as follows: $Q_{tar} = r+\gamma Q(s',\max\limits_{a'}Q(s',a';\theta);\phi)$.

{\bf Deep Deterministic Policy Gradient (DDPG)}:
\vspace{0.2cm}
DDPG is a model-free and off-policy framework which uses a deep neural network for function approximation. But unlike DQN which can only solve discrete and low-dimensional action spaces, DDPG can solve continuous action spaces. In addition, DDPG is an Actor-Critic method that has both a value function network ({\em critic}) and a policy network. The critic network used in DDPG is the same as the actor-critic network described before.  DDPG is derived from DDQN~\cite{van2016deep} and works more robustly by creating two DQNs (target and now) to estimate the value functions.  Thus, this paper adopts the DDPG method to learn concurrently the desired $Q$ function and the corresponding policy.
\vspace{-0.55cm}
%%%%%%%%%%%%%%%%%%%%%%%%%%%%%%%%%%%%%%%%%%%%%%%%%

%%%%%%%%%%%%%%%%%%%%%%%%%%%%%%%%%%%%%%%%%%%%%%%%%

\section{Method}
\label{sec:method}
\vspace{-0.35cm}
% -----------------------------------------------
This paper proposes a cooperative, multi-objective architecture with age-decaying weights for traffic signal control optimization.  It represents each intersection with a DDPG architecture, which contains a critic network and an actor network. The outcome of an action is the number of seconds of green light.  The duration for a phase cycle (green, yellow, red) is different at different intersections but fixed at an intersection.  Also, the duration of yellow light is the same and fixed for all intersections. Then, the duration of the red light can be directly derived once the duration of green is known.     

The original DDPG uses off-policy data and the Bellman equation~\cite{bellmanequation} to learn the $Q$-function, and then derives the policy.  It interleaves learning an approximator to find the best $Q^*(s,a)$ and also learning another approximator to decide the optimal action $a^*(s)$, and so in a way the action space is continuous.  The output of this DDPG is a continuous probability function to represent an action.  In this paper, an action corresponds to the seconds of green light. Although DDPG is off-policy, we  can mix the past data into the training set, making the distribution of the training set diverse by feeding current environment parameters to a traffic simulation platform such as TSIS~\cite{owen2000traffic} or SUMO~\cite{inproceedings} to provide on-policy data for RL training. However, since most of the data provided by other papers use only SUMO for experiments, for fair comparisons, we also use only SUMO for performance evaluations and ablation studies.

In a general DDPG, to increase the opportunities for the agents to explore the environment, random-sampling noise is added to the output action space. However, random perturbations also cause the agent to blindly explore the environment.  In traffic signal control,  most SoTA methods use multiple local agents to model different intersections.   During training, the same training mechanism ``adding noise to the action model'' is used to make each agent explore the environment more.  However, ``increasing the whole throughput'' is the same goal for all local agents.  The learning strategy ``adding noise to action model'' will decrease not only the effectiveness of learning but also the total throughput since blinding exploration will make local agents choose conflict actions to other agents.  This means that a cooperation mechanism should be added to the DDPG method among different local agents to increase the final throughput during the learning process. The main novelty of this paper is to introduce a cooperative learning mechanism with a global agent to avoid local agents blindly exploring environments so that overall throughput and learning effectiveness can be significantly improved.  It is decentralized since the global agent is used only during the training stage. 
\vspace{-0.2cm}
\subsection{Cooperative DPGG Network Architecture}
\vspace{-0.2cm}
Most of the policy-based RL methods~\cite{chu2019multi,nishi2018traffic,mousavi2017traffic} used only local agents to perform RL learning for traffic control.  The requirements of a local agent will easily cause conflicts with other agents and result in the divergence problem during optimization.  In this paper, a COoperative Multi-Objective Multi-Agent DDPG (COMMA-DDPG) framework for optimal traffic signal control is designed, where a local agent controls each intersection, and a global agent cooperates with all local agents. 
Fig.~\ref{fig:proces} shows two architectures of our proposed COMMA DPGG mechanism used in the training and inference stages, respectively. In Fig.~\ref{fig:proces}(a), during training, there is a local agent created at each intersection and a global agent that cooperates with all intersections. 
The global agent optimizes the overall rewards and the local agent observes the traffic status from its corresponding intersection and changes the traffic signal accordingly.  After training, as shown in Fig.~\ref{fig:proces}(b), the global agent is no longer needed. Each local agent can directly change the traffic signal by observing all current traffic statuses from all intersections. 

% \begin{algorithm}[]
% \SetAlgoLined
% Initialize critic network $Q(s,a|\theta^{Q})$ and actor network $\mu(s|\theta^{\mu})$ with random weights $\theta^{Q}$ and $\theta^{\mu}$.\\Initialize target network $Q'$ and $\mu '$\ with weights $\theta^{Q'} \leftarrow \theta^{Q}, \theta^{\mu '} \leftarrow \theta^{\mu}$ and also initialize replay buffer $R$.
  
% \For{t=1, ... ,T}{
%     Clean the replay buffer $\bf B$.\\
%     /* $\bf {B}=(B_1,...,B_m ,...,B_M);$ */ \\
%     /* $B^m$: on-policy data for the $m$th intersection */\\
%     /* Generate on-policy data */ \\ 
%     $\bf B$$=GOD(t)$;  \\
%     \For{episode=1, ..., 400}{
%         \For{m=1,..., M, Global}{
%             \If{$m \neq Global$}{$LAU$($\bf B$,$m$);// Update local agents\\}
%             \If{agent=Global}{$GAU$($\bf B$);// Update the global agent\\}
%             }
%         }
%     }
% \caption{COMMA-DDPG traffic signal control RL algorithm.}
% \label{algo:1}
% \end{algorithm}

 Details of this COMMA DPGG algorithm are described in {\bf Algorithm 1}.  All the algorithms are detailed in the supplementary file. Although the DDPG method is off-policy, we use TSIS~\cite{owen2000traffic} and SUMO~\cite{inproceedings} to collect on-policy data for RL training. Details of the on-policy data collection process are described in the GOD (Generating On-policy Data) algorithm (see {\bf Algorithm 2} in the supplementary file). With the set of on-policy data, the parameters of local and global agents are then updated by the {\bf LAU (Local Agent Updating)} algorithm and {\bf GAU(Global Agent Updating)} algorithm , respectively (also see the supplementary file).  Let $W_{G}^m$ represent the global agent's importance to the  $m$th intersection.  Then, the importance $W_{L}^m$ of the $m$th local agent will be 1-$W_{G}^m$.   
 For the $m$th intersection, its next action will be predicted by the GOD and LAU algorithms, respectively,  via an epsilon greedy exploration scheme.  The output seconds of the global agent and the local agent are compared based on $W_{G}^m$ and $W_{L}^m$. Then, the one with higher importance will be chosen to the output seconds. 
 \vspace{-0.5cm}
 \subsection{Generating On-policy Data}
 \vspace{-0.3cm}
% \begin{algorithm}[]
% \SetAlgoLined 
% /* Run one hour of simulation
% with noise  $\eta$*/\\
% Input: \, \ $t$: timestamp\\ 
% \, \, \, \, \, \, \ $\theta^{\mu}_m$: parameters for the $m$th actor network \\
% \, \, \, \, \, \, \ $\theta^{\mu}_G$: parameters for the global actor network \\
% Output: $\bf B$: on-policy data  \\

% $\beta = 0.95^t$; rate for time decline \\
% \For{m=1, ... , M}{
%  Get $W_{G}^m$ from the global actor network with the parameters  $\theta^{\mu}_G$;\\
%  $W_{G}^m$= $\beta \times W_{G}^m$; $W_{L}^m$=1-$W_{G}^m$;\\
% \For{l=1, ... ,3600}{

%  /* $\epsilon$: the probability of choosing to explore */\\
% /* $\eta_{m}$: noise for epsilon greedy exploration*/\\ 
% $p=$ random(0,1);\\
%  $\eta_{m}=
%   \left\{
% \begin{aligned}
% 0, if\  p \leq \epsilon,\\
% random(-5,5), if\  p > \epsilon ;
% \end{aligned}
% \right.$\

%     $a_{l}^m=
%     \left\{
%     \begin{aligned}
%     \mu(s_{l}|\theta^{\mu}_{m})+\eta_{m}, \;\; \text{if} \; W_{L}^m>W_{G}^m,\\
%     {\bm {\mu}}_{G}(s_{l}|\theta^{\mu}_{G})(m)+\eta_{m}, \;\; \text{if} \; W_{L}^m<W_{G}^m;
%     \end{aligned}
% \right.$

%   Execute $a_{l}^m$ and observe $r_{l}^m, s_{l+1}^m$;\\
%   Store transition $(s_{l}^m,a_{l}^m,r_{l}^m,s_{l+1}^m)$ in $\bf{B_m}$;
% }

%   }
%  $\bf {B}=(B_1,...,B_m ,...,B_M);$\\
%  Return(B);
% \caption{ GOD (Generating On-policy Data)}
% \label{algo:2} 
% \end{algorithm}
During the RL-based training process, before starting, we will perform a one hour simulation to collect data (see {\bf Algorithm 2}) and store them in the replay buffer $\bf B$ based on TSIS or SUMO. Let $\bf{B_m}$ be the set of on-policy data collected for training the $m$-th local agent. Then, $\bf B$ is the union of all $\bf{B_m}$, {\em i.e.}, $\bf B$= $(\bf{B_1},...,\bf{B_m} , ..., \bf{B_M})$.  In the process of interacting with the environment, we will add the epsilon greedy and weight-decayed method to the selection of actions. In particular, the epsilon greedy method will gradually reduce epsilon from 0.9 to 0.1. To avoid training biases, at the $t$th training iteration, a time decay mechanism is adopted to decay $W_{G}^m$ by the ratio $(0.95)^t$.  
\vspace{-0.2cm}
%%%%%%%%%%%%%%%%%%%%%%%%%%%%%%%%%%%%%%%%%%%%%%%%%%%%%%%%%%
\begin{table}[t]  %table 2→1
\scriptsize
%\footnotesize
\setlength\tabcolsep{2.5pt}
\centering
\caption{ Comparisons of throughput among different SoTA methods.
\vspace{-0.3cm}
 }
\begin{tabular}{lcccccc}%lcc代表三欄，第一欄靠左，二三欄置中，兩旁的直線代表欄與欄間劃上直線
\hline  %劃上一條橫線
   Methods / Throughput   &   &   &   &  &  & \\
   & I-1 & I-2 & I-3 & I-4 & I-5 & Average\\\hline
   Fixed & 1530 & 1560 & 1996 & 2288 & 2291 & 1933\\
   MA-DDPG~\cite{Lowe2017MMDDPG}  & 1782 & 1819  &  2098 & 1896 & 2400 & 1999\\
   PPO~\cite{Schulman2017PPO} & 979 & 957 & 1206 & 1517 & 1619 &1255.6\\
   TD3~\cite{Fujimoto2018TD3} & 1370 & 1394 &  1787 & 2070 & 2147 &1753.6\\\hline
   COMMA-DDPG & 2225  & 2310 & 2784& 3052 & 2868 &2647.8\\\hline
\end{tabular}
\label{tab:throughputcomparison}
\vspace{-0.3cm}
\end{table}

% \begin{table}[t]  %table 1→2
% \scriptsize
% % \footnotesize
% \setlength\tabcolsep{0.5pt}
% \centering
% \caption{Comparisons of waiting time at different hours among fixed scheme, MA-DDPG, and our COMMA-DPPAG method.
% \vspace{-0.3cm}
% }
% \label{tab:waitingtime}
% \begin{tabular}{lcccc}%lcc代表三欄，第一欄靠左，二三欄置中，兩旁的直線代表欄與欄間劃上直線

% \hline  %劃上一條橫線
%   Method/Waiting T(s) &  &  &  & \\  
%         &  First 1/4 hour  &  First 1/2 hour &  First 3/4 hour & Total \\\hline
%   Fixed & 93445  & 309620   &  532042 & 750628\\
%   MA-DDPG~\cite{Lowe2017MMDDPG}   & 114907    & 352258  &  578859 & 829713\\\hline
%   COMMA-DDPG & 61380    & 207552 & 390306 & 580782\\\hline
% \end{tabular}
% \vspace{-0.3cm}
% \end{table}
% 還是這邊表1 2換順序  1 看起來可以縮近來都可以

\begin{table}[t] %table 6→4
\tiny
% \footnotesize
\setlength\tabcolsep{3pt}
\centering
\caption{ Performance comparisons among different SoTA methods on two intersections.}
\vspace{-0.1cm}
\begin{tabular}{lcccccc}%lcc代表三欄，第一欄靠左，二三欄置中，兩旁的直線代表欄與欄間劃上直線
\hline  %劃上一條橫線
    Methods & Delay & Speed & Time loss & Travel time & Waiting time\\\hline
    IDQN~\cite{ault2021reinforcement} & 2745.96 & 11.01 & 258.99 & 227.67 & 217.78\\
    IPPO~\cite{ault2020reinforcement} & 2463.69 & 9.62 & 1576.62 & 236.49 & 1538.05\\
    FMA2C~\cite{chu2016large} & 2734.2 & 11.19 & 151.12 & 226.56 & 69.95\\
    MPLight~\cite{Zheng2019RL} & 2712.2 & 11.19 & 158.55 & 226.53 & 73.71\\
    MPLight full(MPLight+IDQN) & 2709.93 & 11.19 & 186.94 & 226.51 & 90.81\\\hline
    COMMA-DDPG & 522 & 14.56 & 138.4 & 156.56 & 38.4 \\\hline
\end{tabular}%\\\\
\label{tab:ComarionsBenchmark}
\vspace{-0.65cm}
\end{table}

\begin{table}[t] 
\tiny
% \footnotesize
\setlength\tabcolsep{3pt}
\centering
\caption{ Performance comparisons among different SoTA methods on five intersections.}
\vspace{-0.1cm}
\begin{tabular}{lcccccc}%lcc代表三欄，第一欄靠左，二三欄置中，兩旁的直線代表欄與欄間劃上直線
\hline  %劃上一條橫線
    Methods & Delay & Speed & Time loss & Travel time & Waiting time\\\hline
    IDQN~\cite{ault2021reinforcement} & 1527.49 & 10.72 & 715.76 & 264.53 & 201.27\\
    IPPO~\cite{ault2020reinforcement} & 1789.1 & 7.27 & 1268.89 & 346.06 & 658.61\\
    FMA2C~\cite{chu2016large} & 1434.68 & 10.72 & 275.69 & 167.96 & 254.54\\
    MPLight~\cite{Zheng2019RL} & 1489.91 & 10.61 & 322.02 & 243.11 & 260.19\\
    MPLight full (MPLight+IDQN) & 1778.32 & 9.32 & 931.36 & 280.76 & 281.14\\\hline
    COMMA-DDPG & 466.05 & 10.75 & 115.74 & 113.25 & 134.89 \\\hline
    % $\vspace{-0.7cm}$
\end{tabular}
\label{tab:ComarionsBenchmark5Intersections}
\end{table}

\begin{table}[t] %table 5
\tiny
% \footnotesize
\setlength\tabcolsep{3pt}
\centering
\caption{ Performance of our method on 16 intersections.}
% \vspace{-0.1cm}
\begin{tabular}{lcccccc}%lcc代表三欄，第一欄靠左，二三欄置中，兩旁的直線代表欄與欄間劃上直線
\hline  %劃上一條橫線
    Methods & Travel time & avg. Waiting time & Speed & Fuel(mg/s) & CO(mg/s) & CO2(mg/s) \\\hline
    Our methods & 1680.89 & 217.51 & 9.2 & 0.93 & 108.53 & 2160.67  \\\hline
    No global agent & 1857.74 & 269.57 & 7.97 & 1.03 & 114.57 & 2208.37 \\\hline
\end{tabular}%\\\\
\label{tab:Performance on 16 intersections}
\vspace{-0.2cm}
\end{table}

%%%%%%%%%%%%%%%%%%%%%%%%%%%%下面移動到附錄 並改成表6
% \begin{table}[t] %table 5
% \tiny
% % \footnotesize
% \setlength\tabcolsep{3pt}
% % \centering
% \caption{ Performance comparisons among different SoTA methods when ten intersections were included.}
% \vspace{-0.1cm}
% \begin{tabular}{lcccccc}%lcc代表三欄，第一欄靠左，二三欄置中，兩旁的直線代表欄與欄間劃上直線
% \hline  %劃上一條橫線
%     Methods & Delay & Speed & Time loss & Travel time & Waiting time\\\hline
%     IDQN~\cite{ault2021reinforcement} & 1814.18 & 10.54 & 771.29 & 517.82 & 669.59\\
%     IPPO~\cite{ault2020reinforcement} & 1861.14 & 9.71 & 1371.08 & 664.55 & 1201.17\\
%     FMA2C~\cite{chu2016large} & 1784.95 & 10.61 & 668.03 & 512.62 & 569.59\\
%     MPLight~\cite{Zheng2019RL} & 1750.39 & 10.69 & 569.07 & 523.38 & 492.83\\
%     MPLight full(MPLight+IDQN) & 1685.54 & 10.91 & 431.79 & 520.78 & 324.46\\\hline
%     COMMA-DDPG & 603.94 & 11.06 & 202.03 & 493.71 & 204.59 \\\hline
% \end{tabular}%\\\\
% \label{tab:ComarionsBenchmark10Intersections}
% \vspace{-0.5cm}
% \end{table}
%%%%%%%%%%%%%%%%%%%%%%%%%%%%%%%%%%%%%%%%%%%%%%%%%%%%%%%%%%
 % \vspace{-0.1cm}
 \subsection{Local Agent}
 \vspace{-0.2cm}
In our scenario, a fixed duration of a traffic signal change cycle is assigned to each intersection.  Furthermore, there are $Y$ seconds prepared for yellow light. Then, we only need to model the phase duration for the green light.  After that, the phase duration of the red light can be directly estimated.  At each intersection,  a DPGG-based architecture is constructed to model its local agent for traffic control.  To describe this local agent, some definitions are given below.
\vspace{-0.3cm}
%\begin{itemize}[leftmargin=16pt]
%\setlength\itemsep{-0.028cm}
\begin{enumerate}
\item The duration of traffic phase ranges from $D_{min}$ to $D_{max}$ seconds.
\item Stopped vehicles are defined as those vehicles whose speeds are less than 3 $km/hr$.
\item The state at an intersection is defined by a vector in which each entry records the number of stopped vehicles of each lane at this intersection at the end of the green light, and current traffic signal phase.
\end {enumerate}
%\end{itemize}
\vspace{-0.3cm}
The reward for evaluating the quality of a state at an intersection is defined as the degree of clearance of this state, {\em i.e.}, the number of vehicles remaining at the intersection when the green-light period ends. There are two cases in which a reward is given to qualify a state; that is, (1) the green light ends, but there are still some vehicles and (2) the green light is still but there is no vehicle.   There is no reward or penalty for other cases.  Let $N_{m,t}$ denote the number of vehicles at intersection $m$ at time $t$, and let $N_{max}$ be the maximum traffic flow.   This paper uses the clearance degree as a reward for qualifying the $m$th local agent.  When the green light ends and there is no vehicle, a pre-defined max reward $R_{max}$ is assigned to the $m$th local agent.  If there are still some vehicles, a penalty proportional to $N_{m,t}$ is assigned to this local agent.  More precisely, for Case 1, the reward $r_{m,t}^{local}$ for the intersection $m$ is defined as: \\
 {\bf Case 1}: If the green light ends but some vehicles are still,
\vspace{-0.2cm}
\begin{equation}
r_{m,t}^{local} = 
\begin{cases}
 R_{max}, if\;\frac{{\;\;{N_{m,t}}}}{{{N_{max}}}} \le \frac{1}{N_{max}}; \\
{-\frac{R_{max}N_{m,t}}{N_{max}}}, \text{else.}\
\end{cases}
\end{equation}
% \vspace{-0.2cm}
For Case 2, if there is no traffic but a long period still remaining for the green light, various vehicles moving on another road should stop and wait until this green light turns off.  To avoid this case, a penalty should be given to this local agent.  Let $g_{m,t}$ denote the remnant green light time (counted by seconds) when there is no traffic flow in the $m$th intersection at time step $t$, and $G_{max}$ the largest duration of green light.  Then, the reward function for Case 2 is defined as:\\
{\bf Case 2}: If there is no traffic but the green light is still on,
$\vspace{-0.1cm}$
\begin{equation}
r_{m,t}^{local} =
\begin{cases}
R_{max}, if\;\frac{{\;\;{g_{m,t}}}}{{{G_{max}}}} \le \frac{1}{G_{max}}; \\
{ -  \frac {R_{max}g_{m,t}} {G_{max}}}, \text{else.}\
\end{cases}
\vspace{-0.1cm}
\end{equation}
% \begin{equation}
%     r_{m,t}^{local} = \left\{ {\begin{array}{*{20}{c}}{{R_{max}},\;\;if\;\frac{{\;\;{g_{m,t}}}}{{{G_{max}}}} \le \frac{1}{G_{max}},\\{ \\{ - R_{max} \times {g_{m,t}}/{G_{max}},\; else.\\\end{array}}}}} \right.
% \end{equation}
% \begin{figure}[t]
%      \centering
%      (a)\includegraphics[scale=0.28]{LaTeX/local_agent_critic.pdf}
%      \centering
%      (b)\includegraphics[scale=0.28]{LaTeX/local_agent_actor.pdf}
%      \caption{Architectures for local agent. (a) Local critic.(b) Local actor.}
%      \label{fig:local_agent}
%  \end{figure}
% \begin{figure}[t]
%      \centering
%      \includegraphics[scale=0.17]{LaTeX/Local_agent_a&b.pdf}
%      \caption{Architectures for local agent. (a) Local critic.(b) Local actor.}
%      \label{fig:local_agent}
%  \end{figure} 
% \vspace{-0.3cm}
Detailed architectures for local agents are shown in the supplementary file (see Fig. 1 in Section C).
% \ref{fig:local_agent}.  
 Its inputs are the number of vehicles stopped at the end of the green light at each lane, the remaining green light seconds, and the current traffic signal phases of all intersections. Thus, the input dimension for each local critic network is $(2M+\sum _{m = 1}^{ {M} }{N_{lane}^m})$, where $M$ denotes the number of intersections and ${N_{lane}^m}$ is the number of lanes in the $m$th intersection.  Then, a hyperbolic tangent function is used as an activation function to normalize all input and output values.  There are two fully connected hidden layers used to model the $Q$-value.  The output is the expected value of future return of doing this action at the state.
% \vspace{-0.1cm}
The architecture of the local actor network is shown in are shown in the supplementary file (see Fig. 1(b) in Section C).
% \ref{fig:local_agent}(b).  
The inputs used to model this network include the numbers of stopped vehicles at the end of the green light at each lane, and current traffic signal phases of all intersections. Thus, the dimension for each local actor network is $(M+\sum _{m = 1}^{ {M} }{N_{lane}^m})$.
Let $\theta ^Q_m$ and $\theta ^{\mu}_m$ denote the sets of parameters of the $m$th local critic and actor networks, respectively. To train $\theta ^{Q}_m$ and $\theta ^{\mu}_m$, we sample a random minibatch of $N_b$ transitions $(\boldsymbol{S}_{i},\boldsymbol{A}_{i},\boldsymbol{R}_{i},\boldsymbol{S}_{i+1})$ from $\bf B$, where  
%\begin{itemize}[leftmargin=16pt]
%\setlength\itemsep{-0.028cm}
\vspace{-0.3cm}
\begin{enumerate}
\item each state $\boldsymbol{S}_{i}$ is an $M\times 1$ vector containing the local states of all intersections;
\item each action  $\boldsymbol{A}_{i}$ is an $M\times 1$ vector containing the seconds of current phase of all intersections; 
\item each reward $\boldsymbol{R}_{i}$ is an $M\times 1$ vector containing the rewards obtained from each intersection after performing $\boldsymbol{A}_i$ at the state $\boldsymbol{S}_i$.  The $m$th entry of $\boldsymbol{R}_{i}$ is the reward of the $m$th intersection after performing $\boldsymbol{A}_i$.
\end{enumerate}
%\end{itemize}

% \begin{algorithm}[]
% \SetAlgoLined
% Input: \\ 
% \ \ \ $\bf B$: on-policy data; $m$: the $m$th agent  \\
% \ \ \ $\theta_m^Q$: set of parameters for the local critic network; \\
% \ \ \ $\theta_m^\mu$: set of parameters for the local actor network;  \\
% \ \ \ ($\theta_m^{Q'}$,$\theta_m^{\mu '}$): sets of parameters for the target network;  \\
% Output: \\ 
% \ \ \ $\theta_m^Q$: new parameters for the $m$th critic network; \\
% \ \ \ $\theta_m^\mu$: new parameters for the $m$th  actor network;  \\
% \ \ \ ($\theta_m^{Q'}$,$\theta_m^{\mu '}$): new parameters for the target network;  \\
% Sample a random minibatch of $N_b$ transitions $({\bm S}_{i},{\bm A}_{i},{\bm R}_{i},{\bm S}_{i+1})$ from $\bf B$;\\
% Set $y_{i}^m={\bm R}_{i}(m)+\gamma Q'({\bm S}_{i+1}|\mu '({\bm S}_{i+1}|\theta_m^{\mu '})|\theta_m^{Q'})$;\\
%   Update the critic parameters $\theta_m^Q$ by minimizing the loss: $L_{critic}^m=\frac{1}{N_b} \sum_{i}(y_{i}^m-Q({\bm S}_{i},{\bm A}_{i}|\theta_m^{Q}))^{2}$;\\
%   Update the actor parameters $\theta_m^\mu$ by minimizing the loss: $L_{actor}^m=-\frac{1}{N_b}\sum_{i}Q({\bm S}_{i},\mu ({\bm S}_{i}|\theta_m^{\mu})|\theta_m^{Q})$;\\
%   Update the target network:\\
%   $\theta_m^{Q'} \leftarrow (1-\tau) \theta_m^{Q}+\tau\theta_m^{Q'};\\
%   \theta_m^{\mu'} \leftarrow (1-\tau) \theta_m^{\mu}+\tau\theta_m^{\mu'}$;
% \caption{LAU (Local Agent Updating)}
% \label{algo:3} 
% \end{algorithm}

\vspace{-0.3cm}
Let $y_i^m$ denote the reward after performing $\boldsymbol{A}_i$ from the $m$th target critic network.  Based on $y_i^m$, the loss functions for updating $\theta^Q_m$ and $\theta^{\mu}_{m}$ are defined, respectively, as follows:
$\vspace{-0.2cm}$
\begin{equation}
L_{critic}^m={\frac{1}{N_b}}\sum_{i=1}^{N_b}(y_{i}^m-Q(\boldsymbol{S}_{i},\boldsymbol{A}_{i}|\theta_m^{Q}))^{2}~~
\mathrm{and} ~~ L_{actor}^m=-{\frac{1}{N_b}}\sum_{i=1}^{N_b}Q(\boldsymbol{S}_{i},\mu (\boldsymbol{S}_{i}|\theta_m^{\mu})|\theta_m^{Q}).
% L_{critic}^m={1 \over {N_b}}\sum_{i=1}^{N_b}(y_{i}^m-Q(\boldsymbol{S}_{i},\boldsymbol{A}_{i}|\theta_m^{Q}))^{2}~~
 % \mathrm{and} ~~ L_{actor}^m=-{1 \over {N_b}}\sum_{i=1}^{N_b}Q(\boldsymbol{S}_{i},\mu (\boldsymbol{S}_{i}|\theta_m^{\mu})|\theta_m^{Q}).
\end{equation}
$\vspace{-0.2cm}$
With $\theta^{Q}_{m}$ and $\theta^{\mu}_{m}$, the parameters $\theta^{Q'}_{m}$ and $\theta^{\mu'}_{m}$ for the target 
network are attentively updated as follows:
% $\theta_m^{Q'} \leftarrow (1-\tau) \theta_m^{Q}+\tau\theta_m^{Q'}$

\begin{equation}
\theta_m^{Q'} \leftarrow (1-\tau) \theta_m^{Q}+\tau\theta_m^{Q'}~~
 \mathrm{and} ~~\theta_m^{\mu'} \leftarrow (1-\tau) \theta_m^{\mu}+\tau\theta_m^{\mu'}.
\end{equation}

The parameter $\tau$ is set to 0.8 for updating the target network. Details to update the parameters of local agents are described in {\bf Algorithm 3} (see the supplementary file).  

To make the output action no longer blindly explore the environment, we introduce a global agent to explore the environment more precisely. The global agent controls the total  waiting time at all intersections. The details of the global critic and actor networks are shown in the supplementary file (Fig.2 in Section C), where (a) is for the global critic network and (b) is for the global actor network. For the $m$th intersection, we use $V_m$ to denote the number of total vehicles, and $T_{m,n}^{w,i}$ to be the waiting time of the $n$th vehicle at the time step $i$. Then, the total waiting time across the whole site is used to define the global reward as follows:~$r_{i}^G=-\frac{1}{M}\sum_{m=1}^{M}\sum_{n=1}^{V_m}T_{m,n}^{w,i}$.  Let $\theta^Q_G$ and $\theta^{\mu}_G$ denote the parameters of the global critic and actor networks, respectively. To train $\theta^Q_G$ and
$\theta^{\mu}_G$, we sample a random minibatch of $N_b$ transitions $(\boldsymbol{S}_{i},\boldsymbol{A}_{i},\boldsymbol{R}_{i},\boldsymbol{S}_{i+1})$ from $\bf B$.  Let $y_i^G$ denote the reward after performing $A_i$ got from the global target critic network.  Then, the loss function for updating $\theta^Q_G$ is defined as follows~:
% $L_{critic}^{G}=\frac{1}{N_b} \sum_{i=1}^{N_b}({y}_{i}^G-{Q_G}({\bm S}_{i},{\bm A}_{i}|\theta^{ Q}_G))^{2}$.\\
$\vspace{-0.3cm}$
\begin{equation}
L_{critic}^{G}=\frac{1}{N_b} \sum_{i=1}^{N_b}({y}_{i}^G-{Q_G}(\boldsymbol{S}_{i},\boldsymbol{A}_{i}|\theta^{ Q}_G))^{2}.
\vspace{-0.2cm}
\end{equation}
$\vspace{0.2cm}$

It is noticed that the output of this global critic network is a
scalar value, {\em i.e.}, the predicted total waiting time across the entire site. To train $\theta^ {\boldsymbol{\mu}_{G}}$, we use the loss function~:
% $L_{actor}^G=- {1 \over {N_b} } \sum_{i=1}^{N_b}{Q_G}({\bm S}_{i},{\bm{\mu_G}} ({\bm S}_{i}|\theta^{\bm{\mu}}_G)|\theta^{Q}_G)$.\\
\begin{equation}
L_{actor}^G=- {\frac{1}{N_b}} \sum_{i=1}^{N_b}{Q_G}(\boldsymbol{S}_{i},\boldsymbol{{\mu_G}} (\boldsymbol{S}_{i}|\theta^{\boldsymbol{\mu}_G})|\theta^{Q}_G).
\end{equation}
In addition, the outputs of the global actor network are an $M\times 1$ vector to output the suggested actions at all intersections, and the weight $W_{G}^m$ to represent the importance of the $m$th intersection of the global agent.   
 All local agents and the global agent are modeled by a DDPG network.  Details to update the global agent are described in {\bf Algorithm 4} (see the supplementary file).  We use the TSIS and SUMO simulation platforms to generate various small or large vehicles moving on the roads through intersections.  
\vspace{-0.2cm}
\subsection{Carbon Emission Reduction}
\vspace{-0.2cm}
%CO的計算如上面的公式，其中參數的部分會在補充資料說明，CO2的公式跟CO是相同的
Another important issue in traffic sign control is to reduce carbon emissions.This paper also adopts the HBEFA formula built in the traffic flow simulation software SUMO for recording and outputing the current vehicle's fuel consumption, carbon emission, and other data in real time since this HBEFA formula is also applicable to some calculations in European countries. The calculation equations of CO are described below, and the parameter part will be explained in the supplementary information, and the content of the formula of CO2 is similar to that of CO, only CO is replaced by CO2.
\begin{equation}
CO_{move} = (CO_{engine}*V_{engine}*FC*M_{fuel})/(M_{air}*1000),
\label{eq:CO}
\end{equation}
\begin{equation}
CO_{stop} = (CO_{engine}*V_{engine}*r_{stop}*t_{stop})/(3600*M_{air}),
\label{eq:CO_stop}
\end{equation}
\begin{equation}
CO = CO_{move}+CO_{stop}.
\label{eq:CO=CO+CO_stop}
\end{equation}

We can see from the formula in HBEFA that the main influencing factors are the distance traveled (v) and the waiting time (${t_{stop}}$). Distance affects the performance of the fuel cell (FC) and will remain fixed during our experiment, so the influence of waiting time is the main source of difference. By examining Eqs. (\ref{eq:CO}) and (\ref{eq:CO_stop}), clearly, if ${t_{stop}}$ is reduced, carbon emissions are also reduced.
  
\vspace{-0.2cm}

\section{Experimental Results}
\label{sec:results}
\vspace{-0.2cm}
Our traffic data consist of visual traffic monitoring sequences from five consecutive intersections during the morning rush hour in a midsize city in Asia. In order to facilitate comparison with other SOTA papers, we used SUMO traffic simulation software for simulation.We take a fixed-time traffic light control scheme with one hour total waiting time as a baseline for comparisons. We also performed ablation studies on the COMMA-DDPG approach with and without the global agent to make comparisons.  In addition, we also evaluated an open benchmark~\cite{ault2021reinforcement} to make fair comparisons with other SoTA methods.
%  \begin{figure}[t]
%      \centering
%      \includegraphics[scale = 0.4]{LaTeX/waitingtime.png}
%      \caption{Waiting time converge conditions during the training process among different methods.}
%      \label{fig:WaitingTime}
%  \end{figure}
%%%%%%%%%%%%%%%%%%%%%%%%%%%%%%%%%%%%%%%%%%%%%%%%%

To train our model, we first used the fixed-time control model to pretrain our COMMA-DDPG model. Fig. 4 (in the supplement) shows the converge conditions of waiting time among different methods.  Clearly, the fixed-time control model performs better than the MA-DDPG model.  However, they did not converge.  As to our proposed COMMA-DDPG method, it gradually and robustly converges to a local minima which is the best than the other two methods.  
% Table~\ref{tab:waitingtime} shows the comparisons of waiting time at different hours among them.   Clearly, the waiting time got of our method along time change is much shorter than the ones got by the fixed-time model and the MA-DDPG method, respectively.
Table~\ref{tab:throughputcomparison} shows the throughtput comparisons among the fixed-time, MA-DDPG, PPO, TD3, and our COMMA-DDPG schemes at the five observed intersections.  Due to the global agent, the throughput obtained by our method is much higher than the baseline and the MA-DDPG method.  When training agents over the larger road network (more than 10 intersections), it will be problematic that both local and global agents take the information of all intersections as the input.  To reflect real situations, we adjust the global agent by taking only 8 nearby intersections (up, down, left, right, upper left, lower left, upper right, lower right) as a training basis.  This means that the local agent can still contain global information. 

To make fair comparisons with other SoTA methods, there is an RL testbed environment for traffic signal control~\cite{ault2021reinforcement}.  It is based on the well-established Simulation of Urban Mobility traffic simulator (SUMO).  It includes single- and multiagent-signal control tasks that are based on realistic traffic scenarios from SUMO.  To allow easy easy deployment of standard RL algorithms, an OpenAI GYM interface is also provided.  It also provides open source data and codes of SoTA RL-based signal control algorithms for performance evaluation.  There are five SoTA methods provided for performance evaluations; that is, IDQN~\cite{ault2021reinforcement}, IPPO~\cite{ault2020reinforcement}, FMA2C~\cite{chu2016large}, MPLight~\cite{Zheng2019RL}, and MPLight full~\cite{ault2021reinforcement,Zheng2019RL}. MPLight is a phase competition modeling method. IDQN and IPPO are decentralized algorithms that can effectively learn instance dependent features.  FMA2C is a large-scale multi-agent reinforcement learning method for traffic signal control. ``MPLight full'' is similar to the MPLight implementation but sensing information matched with IDQN appended to the existing pressure state.  The state and reward functions were set according to the definitions of each algorithm.   Five reward metrics adopted in this paper for performance comparisons are: delay, speed, time loss, system travel time, and total waiting time at intersections.  Table~\ref{tab:ComarionsBenchmark} shows the performance comparisons among these methods and our COMMA-DDPG scheme when only two intersections were included. IDQNN~\cite{ault2021reinforcement} is a set of independent DQN agents, one per intersection, each with convolution-layers for lane aggregation.  IPPO has the same deep neural network of IDQN but the output layer which is constructed with a set of polynomial functions. IPPO performs better than IDQNN in ``delay'' but much worse in waiting time due to the under-fitting problem of the set of polynomial functions. IDQNN and IPPO are formed by independent DQN agents and thus perform worse than multi-agent methods such as FMA2C and MPLight in the ``time loss'', ``travel time'', and ``Waiting time'' categories. MPLight~\cite{Zheng2019RL} is a decentralized deep reinforcement learning method that uses the concept of pressure to coordinate multiple intersections. It outperforms IDQN~\cite{ault2021reinforcement} and IPPO~\cite{ault2020reinforcement} in the categories ``speed'', ``time loss'', ''travel time'', and ``waiting time'' categories.  It performs worse than ``FMA2C'' and our method.  FMA2C~\cite{chu2016large} is a multi-agent RL method that overcomes the scalability issue by distributing the global control to each local RL agent. It uses a hierarchy of managing agents to enable cooperation between
signal control agents (one per intersection).  However, as described in~\cite{ault2021reinforcement}, it requires much more training episodes to converge than IDQN and
MPLight. It outperforms all the other methods (IDQN, IPPO, MPLight, MPLight FUll) in all metrics. As to our COMMA-DDPG method, it includes a global agent to train each local agent to make better actions during the training stage.  Since the global agent is not included during inference, it also is a decentralized RL-based method for traffic signal control.  With the help of the global agent, each local agent in our COMMA-DDPG architecture can choose non-conflict actions to other agents for better traffic signal control.  Clearly, our COMMA-DDPG method outperforms all SoTA methods in all categories.  It is noticed that our method has impressive results in the ``'Delay', ``Travel time'', and ``Waiting time'' categories. 

When more intersections are added, the stability and generality of our method can be proved. Table~\ref{tab:ComarionsBenchmark5Intersections} shows the performance comparisons among different SoTA methods when five intersections were included to build the road networks. In this case, IPPO still shows significant instability and performs worse in almost performance metrics, especially in ``time loss''.
MPLight-full performs better than the IPPO method. IDQN~\cite{ault2021reinforcement} performs better at ``speed'' and ``waiting time''.  FMA2C outperforms other methods in many performance metrics such as ``Delay'', ``Speed'', ``Travel time'', and ``Waiting time'' but still performs worse than our method.  However, even though five intersections are added, our COMMA-DPPG method still outperforms all SoTA methods. In Table 5, we show data for 16 intersection conditions, including travel time, average, waiting time, speed, CO, CO2, and fuel. The design of this map is taken from real life, bringing together several larger junctions into a 4$\times$4 checkerboard map.  Finally, we can see that our method has better performance than that without the global agent, and according to the HBEFA formula built in SUMO, it can be seen that in terms of environmental protection, CO2, etc. have also been reduced.

\vspace{-0.2cm}
\section{Conclusion}
\vspace{-0.2cm}
This paper proposed a novel cooperative RL architecture to
handle cooperation problems by adding a global agent. Since
the global agent knows all the intersection information, it can guide the local agent to make better actions in the training process. Thus, the local agent does not need to use random noise to randomly explore the environment. Since RL training requires a large amount of data, we hope to add it to RL through data augmentation in the future, so that training can be more efficient. The weakness of our method is that all information of local agents need to be sent to other agents.  In the near future, the COMMA-DPPG will be really evaluated in real road conditions.  
\vspace{-0.2cm}
\section{Appendix for Convergence Proof}
\vspace{-0.2cm}
% Details of the proof of convergence are provided in the supplementary file, with experimental data for an additional 10 junctions.
We additionally put the proof and the experimental data of more intersections in the appendix.

% \clearpage
{\small
\bibliographystyle{unsrtnat}
\bibliography{egbib}

\begin{thebibliography}{44}
\providecommand{\natexlab}[1]{#1}
\providecommand{\url}[1]{\texttt{#1}}
\expandafter\ifx\csname urlstyle\endcsname\relax
  \providecommand{\doi}[1]{doi: #1}\else
  \providecommand{\doi}{doi: \begingroup \urlstyle{rm}\Url}\fi

\bibitem[Alemzadeh et~al.(2020)Alemzadeh, Moslemi, Sharma, and
  Mesbahi]{alemzadeh2020adaptive}
Siavash Alemzadeh, Ramin Moslemi, Ratnesh Sharma, and Mehran Mesbahi.
\newblock Adaptive traffic control with deep reinforcement learning: Towards
  state-of-the-art and beyond.
\newblock \emph{arXiv preprint arXiv:2007.10960}, 2020.

\bibitem[Zheng et~al.(2019{\natexlab{a}})Zheng, Zang, Xu, Wei, Yu, Gayah, Xu,
  and Li]{zheng2019diagnosing}
Guanjie Zheng, Xinshi Zang, Nan Xu, Hua Wei, Zhengyao Yu, Vikash Gayah, Kai Xu,
  and Zhenhui Li.
\newblock Diagnosing reinforcement learning for traffic signal control.
\newblock \emph{arXiv preprint arXiv:1905.04716}, 2019{\natexlab{a}}.

\bibitem[Wei et~al.(2021)Wei, Zheng, Gayah, and Li]{wei2021recent}
Hua Wei, Guanjie Zheng, Vikash Gayah, and Zhenhui Li.
\newblock Recent advances in reinforcement learning for traffic signal control:
  A survey of models and evaluation.
\newblock \emph{ACM SIGKDD Explorations Newsletter}, 22\penalty0 (2):\penalty0
  12--18, 2021.

\bibitem[Mannion et~al.(2016)Mannion, Duggan, and
  Howley]{mannion2016experimental}
Patrick Mannion, Jim Duggan, and Enda Howley.
\newblock An experimental review of reinforcement learning algorithms for
  adaptive traffic signal control.
\newblock \emph{Autonomic road transport support systems}, pages 47--66, 2016.

\bibitem[Pham et~al.(2013)Pham, Brys, Taylor, Brys, Drugan, Bosman, Cock,
  Lazar, Demarchi, Steenhoff, et~al.]{pham2013learning}
Tong~Thanh Pham, Tim Brys, Matthew~E Taylor, Tim Brys, Madalina~M Drugan,
  PA~Bosman, Martine-De Cock, Cosmin Lazar, L~Demarchi, David Steenhoff, et~al.
\newblock Learning coordinated traffic light control.
\newblock In \emph{Proceedings of the Adaptive and Learning Agents workshop (at
  AAMAS-13)}, volume~10, pages 1196--1201. IEEE, 2013.

\bibitem[Van~der Pol and Oliehoek(2016)]{van2016coordinated}
Elise Van~der Pol and Frans~A Oliehoek.
\newblock Coordinated deep reinforcement learners for traffic light control.
\newblock \emph{Proceedings of Learning, Inference and Control of Multi-Agent
  Systems (at NIPS 2016)}, 2016.

\bibitem[Wei et~al.(2018)Wei, Zheng, Yao, and Li]{wei2018intellilight}
Hua Wei, Guanjie Zheng, Huaxiu Yao, and Zhenhui Li.
\newblock Intellilight: A reinforcement learning approach for intelligent
  traffic light control.
\newblock In \emph{Proceedings of the 24th ACM SIGKDD International Conference
  on Knowledge Discovery \& Data Mining}, pages 2496--2505, 2018.

\bibitem[Arel et~al.(2010)Arel, Liu, Urbanik, and Kohls]{arel2010reinforcement}
Itamar Arel, Cong Liu, Tom Urbanik, and Airton~G Kohls.
\newblock Reinforcement learning-based multi-agent system for network traffic
  signal control.
\newblock \emph{IET Intelligent Transport Systems}, 4\penalty0 (2):\penalty0
  128--135, 2010.

\bibitem[Calvo and Dusparic(2018)]{calvo2018heterogeneous}
Jeancarlo~Arguello Calvo and Ivana Dusparic.
\newblock Heterogeneous multi-agent deep reinforcement learning for traffic
  lights control.
\newblock In \emph{AICS}, pages 2--13, 2018.

\bibitem[Lutter et~al.(2021)Lutter, Mannor, Peters, Fox, and
  Garg]{ValueIter:Continuous:ICML2021}
Michael Lutter, Shie Mannor, Jan Peters, Dieter Fox, and Animesh Garg.
\newblock Value iteration in continuous actions, states and time.
\newblock In \emph{ICML}, 2021.

\bibitem[Chu et~al.(2019)Chu, Wang, Codec{\`a}, and Li]{chu2019multi}
Tianshu Chu, Jie Wang, Lara Codec{\`a}, and Zhaojian Li.
\newblock Multi-agent deep reinforcement learning for large-scale traffic
  signal control.
\newblock \emph{IEEE Transactions on Intelligent Transportation Systems},
  21\penalty0 (3):\penalty0 1086--1095, 2019.

\bibitem[Nishi et~al.(2018)Nishi, Otaki, Hayakawa, and
  Yoshimura]{nishi2018traffic}
Tomoki Nishi, Keisuke Otaki, Keiichiro Hayakawa, and Takayoshi Yoshimura.
\newblock Traffic signal control based on reinforcement learning with graph
  convolutional neural nets.
\newblock In \emph{2018 21st International Conference on Intelligent
  Transportation Systems (ITSC)}, pages 877--883. IEEE, 2018.

\bibitem[Mousavi et~al.(2017)Mousavi, Schukat, and Howley]{mousavi2017traffic}
Seyed~Sajad Mousavi, Michael Schukat, and Enda Howley.
\newblock Traffic light control using deep policy-gradient and
  value-function-based reinforcement learning.
\newblock \emph{IET Intelligent Transport Systems}, 11\penalty0 (7):\penalty0
  417--423, 2017.

\bibitem[Lillicrap et~al.(2015)Lillicrap, Hunt, Pritzel, Heess, Erez, Tassa,
  Silver, and Wierstra]{lillicrap2015continuous}
Timothy~P Lillicrap, Jonathan~J Hunt, Alexander Pritzel, Nicolas Heess, Tom
  Erez, Yuval Tassa, David Silver, and Daan Wierstra.
\newblock Continuous control with deep reinforcement learning.
\newblock \emph{arXiv preprint arXiv:1509.02971}, 2015.

\bibitem[Pang and Gao(2019)]{pang2019deep}
Hali Pang and Weilong Gao.
\newblock Deep deterministic policy gradient for traffic signal control of
  single intersection.
\newblock In \emph{2019 Chinese Control And Decision Conference (CCDC)}, pages
  5861--5866. IEEE, 2019.

\bibitem[Wu(2020)]{wu2020control}
Haosheng Wu.
\newblock Control method of traffic signal lights based on ddpg reinforcement
  learning.
\newblock In \emph{Journal of Physics: Conference Series}, volume 1646, page
  012077. IOP Publishing, 2020.

\bibitem[Aslani et~al.(2017)Aslani, Mesgari, and Wiering]{aslani2017adaptive}
Mohammad Aslani, Mohammad~Saadi Mesgari, and Marco Wiering.
\newblock Adaptive traffic signal control with actor-critic methods in a
  real-world traffic network with different traffic disruption events.
\newblock \emph{Transportation Research Part C: Emerging Technologies},
  85:\penalty0 732--752, 2017.

\bibitem[Aslani et~al.(2018)Aslani, Seipel, Mesgari, and
  Wiering]{aslani2018traffic}
Mohammad Aslani, Stefan Seipel, Mohammad~Saadi Mesgari, and Marco Wiering.
\newblock Traffic signal optimization through discrete and continuous
  reinforcement learning with robustness analysis in downtown tehran.
\newblock \emph{Advanced Engineering Informatics}, 38:\penalty0 639--655, 2018.

\bibitem[Ault and Sharon(2021)]{ault2021reinforcement}
James Ault and Guni Sharon.
\newblock Reinforcement learning benchmarks for traffic signal control.
\newblock In \emph{Thirty-fifth Conference on Neural Information Processing
  Systems Datasets and Benchmarks Track (Round 1)}, 2021.

\bibitem[Roess et~al.(2004)Roess, Prassas, and McShane]{roess2004traffic}
Roger~P Roess, Elena~S Prassas, and William~R McShane.
\newblock \emph{Traffic engineering}.
\newblock Pearson/Prentice Hall, 2004.

\bibitem[Fellendorf(1994)]{fellendorf1994vissim}
Martin Fellendorf.
\newblock Vissim: A microscopic simulation tool to evaluate actuated signal
  control including bus priority.
\newblock In \emph{64th Institute of Transportation Engineers Annual Meeting},
  volume~32, pages 1--9. Springer, 1994.

\bibitem[Mirchandani and Head(2001)]{mirchandani2001real}
Pitu Mirchandani and Larry Head.
\newblock A real-time traffic signal control system: architecture, algorithms,
  and analysis.
\newblock \emph{Transportation Research Part C: Emerging Technologies},
  9\penalty0 (6):\penalty0 415--432, 2001.

\bibitem[Zheng et~al.(2019{\natexlab{b}})Zheng, Xiong, Zang, Feng, Wei, Zhang,
  Li, Xu, and Li]{zheng2019learning}
Guanjie Zheng, Yuanhao Xiong, Xinshi Zang, Jie Feng, Hua Wei, Huichu Zhang,
  Yong Li, Kai Xu, and Zhenhui Li.
\newblock Learning phase competition for traffic signal control.
\newblock In \emph{Proceedings of the 28th ACM International Conference on
  Information and Knowledge Management}, pages 1963--1972, 2019{\natexlab{b}}.

\bibitem[Lowrie(1990)]{lowrie1990scats}
PR~Lowrie.
\newblock \emph{Scats, sydney co-ordinated adaptive traffic system: A traffic
  responsive method of controlling urban traffic}.
\newblock Darlinghurst, NSW, Australia, 1990.

\bibitem[Watkins and Dayan(1992)]{watkins1992q}
Christopher~JCH Watkins and Peter Dayan.
\newblock Q-learning.
\newblock \emph{Machine learning}, 8\penalty0 (3-4):\penalty0 279--292, 1992.

\bibitem[Abdoos et~al.(2011)Abdoos, Mozayani, and Bazzan]{abdoos2011traffic}
Monireh Abdoos, Nasser Mozayani, and Ana~LC Bazzan.
\newblock Traffic light control in non-stationary environments based on multi
  agent q-learning.
\newblock In \emph{2011 14th International IEEE conference on intelligent
  transportation systems (ITSC)}, pages 1580--1585. IEEE, 2011.

\bibitem[Wei et~al.(2019{\natexlab{a}})Wei, Chen, Zheng, Wu, Gayah, Xu, and
  Li]{wei2019presslight}
Hua Wei, Chacha Chen, Guanjie Zheng, Kan Wu, Vikash Gayah, Kai Xu, and Zhenhui
  Li.
\newblock Presslight: Learning max pressure control to coordinate traffic
  signals in arterial network.
\newblock In \emph{Proceedings of the 25th ACM SIGKDD International Conference
  on Knowledge Discovery \& Data Mining}, pages 1290--1298, 2019{\natexlab{a}}.

\bibitem[Wei et~al.(2019{\natexlab{b}})Wei, Xu, Zhang, Zheng, Zang, Chen,
  Zhang, Zhu, Xu, and Li]{wei2019colight}
Hua Wei, Nan Xu, Huichu Zhang, Guanjie Zheng, Xinshi Zang, Chacha Chen, Weinan
  Zhang, Yanmin Zhu, Kai Xu, and Zhenhui Li.
\newblock Colight: Learning network-level cooperation for traffic signal
  control.
\newblock In \emph{Proceedings of the 28th ACM International Conference on
  Information and Knowledge Management}, pages 1913--1922, 2019{\natexlab{b}}.

\bibitem[Guo et~al.(2014)Guo, Singh, Lee, Lewis, and Wang]{guo2014deep}
Xiaoxiao Guo, Satinder Singh, Honglak Lee, Richard~L Lewis, and Xiaoshi Wang.
\newblock Deep learning for real-time atari game play using offline monte-carlo
  tree search planning.
\newblock \emph{Advances in neural information processing systems},
  27:\penalty0 3338--3346, 2014.

\bibitem[Van~Hasselt et~al.(2016)Van~Hasselt, Guez, and Silver]{van2016deep}
Hado Van~Hasselt, Arthur Guez, and David Silver.
\newblock Deep reinforcement learning with double {Q}-learning.
\newblock \emph{Proceedings of the AAAI Conference on Artificial Intelligence},
  30\penalty0 (1), 2016.

\bibitem[Casas(2017)]{casas2017deep}
Noe Casas.
\newblock Deep deterministic policy gradient for urban traffic light control.
\newblock \emph{arXiv preprint arXiv:1703.09035}, 2017.

\bibitem[Gupta et~al.(2017)Gupta, Egorov, and
  Kochenderfer]{gupta2017cooperative}
Jayesh~K Gupta, Maxim Egorov, and Mykel Kochenderfer.
\newblock Cooperative multi-agent control using deep reinforcement learning.
\newblock In \emph{International Conference on Autonomous Agents and Multiagent
  Systems}, pages 66--83. Springer, 2017.

\bibitem[Lowe et~al.(2017)Lowe, Wu, Tamar, Harb, Abbeel, and
  Mordatch]{Lowe2017MMDDPG}
Ryan Lowe, Yi~Wu, Aviv Tamar, Jean Harb, Pieter Abbeel, and Igor Mordatch.
\newblock Multi-agent actor-critic for mixed cooperative-competitive
  environments.
\newblock In \emph{Proceedings of the 31st International Conference on Neural
  Information Processing Systems}, 2017.

\bibitem[Zhang et~al.(2019)Zhang, Feng, Liu, Ding, and Zhu]{Zhang2019CityFlow}
Huichu Zhang, Siyuan Feng, Chang Liu, Yaoyao Ding, and Yichen Zhu.
\newblock Cityflow: a multi-agent reinforcement learning environment for large
  scale city traffic scenario.
\newblock In \emph{WWW '19: The World Wide Web Conference}, pages 3620--3624,
  2019.

\bibitem[Barron and Ishii(1989)]{bellmanequation}
E.~N. Barron and H~Ishii.
\newblock The bellman equation for minimizing the maximum cost.
\newblock \emph{Nonlinear Analysis: Theory, Methods and Applications},
  13\penalty0 (9):\penalty0 1067--1090, 1989.

\bibitem[Hester et~al.(2018)Hester, Vecerik, Pietquin, Lanctot, Schaul, Piot,
  Horgan, Quan, Sendonaris, Osband, et~al.]{hester2018deep}
Todd Hester, Matej Vecerik, Olivier Pietquin, Marc Lanctot, Tom Schaul, Bilal
  Piot, Dan Horgan, John Quan, Andrew Sendonaris, Ian Osband, et~al.
\newblock Deep {Q}-learning from demonstrations.
\newblock In \emph{Proceedings of the AAAI Conference on Artificial
  Intelligence}, 2018.

\bibitem[Mnih et~al.(2015)Mnih, Kavukcuoglu, Silver, Rusu, Veness, Bellemare,
  Graves, Riedmiller, Fidjeland, Ostrovski, et~al.]{mnih2015human}
Volodymyr Mnih, Koray Kavukcuoglu, David Silver, Andrei~A Rusu, Joel Veness,
  Marc~G Bellemare, Alex Graves, Martin Riedmiller, Andreas~K Fidjeland, Georg
  Ostrovski, et~al.
\newblock Human-level control through deep reinforcement learning.
\newblock \emph{nature}, 518\penalty0 (7540):\penalty0 529--533, 2015.

\bibitem[Owen et~al.(2000)Owen, Zhang, Rao, and McHale]{owen2000traffic}
Larry~E Owen, Yunlong Zhang, Lei Rao, and Gene McHale.
\newblock Traffic flow simulation using corsim.
\newblock In \emph{2000 Winter Simulation Conference Proceedings (Cat. No.
  00CH37165)}, volume~2, pages 1143--1147. IEEE, 2000.

\bibitem[Krajzewicz et~al.(2002)Krajzewicz, Hertkorn, Feld, and
  Wagner]{inproceedings}
Daniel Krajzewicz, Georg Hertkorn, Christian Feld, and Peter Wagner.
\newblock Sumo (simulation of urban mobility); an open-source traffic
  simulation.
\newblock pages 183--187, 01 2002.
\newblock ISBN 90-77039-09-0.

\bibitem[Schulman et~al.(2017)Schulman, Wolski, Dhariwal, Radford, and
  Klimov]{Schulman2017PPO}
John Schulman, Filip Wolski, Prafulla Dhariwal, Alec Radford, and Oleg Klimov.
\newblock Proximal policy optimization algorithms.
\newblock \emph{arXiv preprint arXiv:1707.06347}, 2017.

\bibitem[Fujimoto et~al.(2018)Fujimoto, Hoof, and Meger]{Fujimoto2018TD3}
Scott Fujimoto, Herke~van Hoof, and David Meger.
\newblock Addressing function approximation error in actor-critic methods.
\newblock In \emph{International Conference on Machine Learning}, pages
  1587--1596, 2018.

\bibitem[Ault and Sharon(2020)]{ault2020reinforcement}
James Ault and Guni Sharon.
\newblock Learning an interpretable traffic signal control policy.
\newblock In \emph{Proceedings of the 19th International Conference on
  Autonomous Agents and MultiAgent Systems}, 2020.

\bibitem[Chu et~al.(2016)Chu, Qu, and Wang]{chu2016large}
Tianshu Chu, Shuhui Qu, and Jie Wang.
\newblock Large-scale multi-agent reinforcement learning using image-based
  state representation.
\newblock In \emph{2016 IEEE 55th Conference on Decision and Control (CDC)},
  pages 7592--7597. IEEE, 2016.

\bibitem[Zheng et~al.(2019{\natexlab{c}})Zheng, Xiong, Zhang, Wei, Zhang, Li,
  Xu, and Li]{Zheng2019RL}
Guanjie Zheng, Yuanhao Xiong, Xinshi Zhang, Hua Wei, Huichu Zhang, Yong Li, Kai
  Xu, and Zhenhui Li.
\newblock Learning traffic signal control from demonstrations.
\newblock In \emph{Proceedings of the 28th ACM International Conference on
  Information and Knowledge Management}, pages 2289--2292, 2019{\natexlab{c}}.

\end{thebibliography}
}

%%%%%%%%%%%%%%%%%%%%%%%%%%%%%%%%%%%%%%%%%%%%%%%%%%%%%%%%%%%%%%%%%%%%%%%%%%%%%%%%%%

\section{Supplement}
\subsection{Appendix for Convergence Proof}
In this section, we will prove that value function in our method will actually converge.

\newtheorem{definition}{Definition}
\newtheorem{theorem}{Theorem}
\newtheorem{lemma}{Lemma}
\newtheorem{proof}{Proof}

\begin{definition}
A metric space $<M,d>$ is complete (or Cauchy) if and only if all Cauchy sequences in $M$ will converge to $M$. In other words, in a complete metric space, for any point sequence $a_{1},a_{2}, \cdots \in M$, if the sequence is Cauchy, then the sequence converges to $M$:
\[ \lim_{n \rightarrow \infty}a_{n} \in M. \]
\end{definition}

\begin{definition}
Let (X,d) be a complete metric space. Then, a map T : X $\rightarrow$ X is called a contraction mapping on X if there exists q $\in [0, 1)$ such that $d(T(x),T(y))<qd(x,y)$, $\forall x,y \in X$.
\end{definition}

\begin{theorem}[Banach fixed-point theorem]
Let (X,d) be a non-empty complete metric space with a contraction mapping T : X $\rightarrow$ X. Then T admits a unique fixed-point $x^{*}$ in X. i.e. $T(x^{*})=x^{*}.$
\end{theorem}

\begin{theorem}[Gershgorin circle theorem]
Let A be a complex $n\times n$ matrix, with entries $a_{ij}$. For $i \in {1,2,...,n}$, let $R_{i}$ be the sum of the absolute of values of the non-diagonal entries in the $i^{th}$ row:
$$R_{i}=\sum_{j=0,j\neq i}^{n}|a_{ij}|.$$
Let $D(a_{ii},R_{i})\subseteq \mathbb{C} $ be a closed disc centered at $a_{ii}$ with radius $R_{i}$, and every eigenvalue of ${\displaystyle A}$ lies within at least one of the Gershgorin discs ${\displaystyle D(a_{ii},R_{i}).}$
\end{theorem}
\begin{lemma}
We claim that the value function of RL can actually converge, and we also apply it to traffic control.
\end{lemma}
\begin{proof}
The value function is to calculate the value of each state, which is defined as follows:
\begin{equation}
\begin{array}{l}
{V^\pi }(s) = \sum\limits_a \pi  (a|s)\sum\limits_{s',r} p (s',r|s,a)[r + \gamma {V^\pi }(s')]\\
 = \sum\limits_a \pi  (a|s)\sum\limits_{s',r} p (s',r|s,a)r\\
{\rm{ }} + \sum\limits_a \pi  (a|s)\sum\limits_{s',r} p (s',r|s,a)[\gamma {V^\pi }(s')].
\end{array}
\end{equation}
Since the immediate reward is determined, it can be regarded as a constant term relative to the second term. Assuming that the state is finite, we express the state value function in matrix form below.
Set the state set $S=\{S_{0},S_{1},\cdots,S_{n}\}$, $V^{\pi}=\{ V^{\pi}(s_{0}), V^{\pi}(s_{1}), \cdots , V^{\pi}(s_{n}) \}^{T}$, and the transition matrix is
\begin{equation}
    {P^\pi } = \left( {\begin{array}{*{20}{c}}
0&{P^\pi _{0,1}}& \cdots &{P^\pi _{0,n}}\\
{P^\pi _{1,0}}&0& \cdots &{P^\pi _{1,n}}\\
 \cdots & \cdots & \cdots & \cdots \\
{P^\pi _{n,0}}&{P^\pi _{n,1}}& \cdots &0
\end{array}} \right),
\end{equation}
where $P^\pi _{i,j} = \sum\limits_a {\pi (a|{s_i})p({s_j},r|{s_i},a)}$.  The constant term is expressed as $R^{\pi}=\{ R_{0}, R_{1}, \cdots, R_{n}\}^{T}$. Then we can rewrite the state-value function as:
\begin{equation}
V^{\pi}=R^{\pi}+\lambda P^{\pi}V^{\pi}.
\end{equation}
Above we define the state value function vector as $V^{\pi}=\{ V^{\pi}(s_{0}), V^{\pi}(s_{1}),\cdots, V^{\pi}(s_{n})\}^{T}$, which belongs to the value function space $V$. We consider $V$ to be an n-dimensional vector full space, and define the metric of this space is the infinite norm. It means:
\begin{equation}
    d(u,v)=\parallel u-v \parallel_{\infty}=\max_{s \in S}|u(s)-v(s)|,\forall u,v \in V
\end{equation}
Since $<V,d>$ is the full space of vectors, $V$ is a complete metric space. Then, the iteration result of the state value function is $u_{new}=T^{\pi}(u)=R^{\pi}+\lambda P^{\pi}u$.
We can show that it is a contraction mapping.
\begin{equation}
    \begin{aligned}
    d(T^{\pi}(u),T^{\pi}(v))&=\parallel (R^{\pi}+\lambda P^{\pi}u)-(R^{\pi}+\lambda P^{\pi}v)\parallel_{\infty}\\
    &=\parallel \lambda P^{\pi}(u-v) \parallel_{\infty}\\
    &\le \parallel \lambda P^{\pi}\parallel u-v \parallel_{\infty}\parallel _{\infty}.
    \end{aligned}
\end{equation}
From Theorem 2, we can show that every eigenvalue of $P^{\pi}$ is in the disc centered at $(0,0)$ with radius 1. That is, the maximum absolute value of eigenvalue will be less than 1.
\begin{equation}
    \begin{aligned}
    d(T^{\pi}(u),T^{\pi}(v))&\le \parallel\lambda P^{\pi}\parallel u-v \parallel_{\infty}\parallel_{\infty}\\
    &\le \lambda \parallel u-v \parallel_{\infty}\\
    &=\lambda d(u,v).
    \end{aligned}
\end{equation}
From the Theorem 1, Eq.(2) converges to only $V^{\pi}$.
\end{proof}
%%%%%%%%%%%%%%%%%%%%%%%%%%%%%%%%%%%%%%%%%%%%%%%%%%%%%%%%%%%%%%%%%%%%%%%%%%%%%%%
% \newpage
% \appendix
\subsection{Algorithm}
% \vspace{-1cm}
% \begin{algorithm}[H]
% \SetAlgoLined
% Initialize critic network $Q(s,a|\theta^{Q})$ and actor network $\mu(s|\theta^{\mu})$ with random weights $\theta^{Q}$ and $\theta^{\mu}$.\\Initialize target network $Q'$ and $\mu '$\ with weights $\theta^{Q'} \leftarrow \theta^{Q}, \theta^{\mu '} \leftarrow \theta^{\mu}$ and also initialize replay buffer $R$.
  
% \For{t=1, ... ,T}{
%     Clean the replay buffer $\bf B$.\\
%     /* $\bf {B}=(B_1,...,B_m ,...,B_M);$ */ \\
%     /* $B^m$: on-policy data for the $m$th intersection */\\
%     /* Generate on-policy data */ \\ 
%     $\bf B$$=GOD(t)$;  \\
%     \For{episode=1, ..., 400}{
%         \For{m=1,..., M, Global}{
%             \If{$m \neq Global$}{$LAU$($\bf B$,$m$);// Update local agents\\}
%             \If{agent=Global}{$GAU$($\bf B$);// Update the global agent\\}
%             }
%         }
%     }

\begin{figure}[H]
\centering
\includegraphics[scale = 0.55]{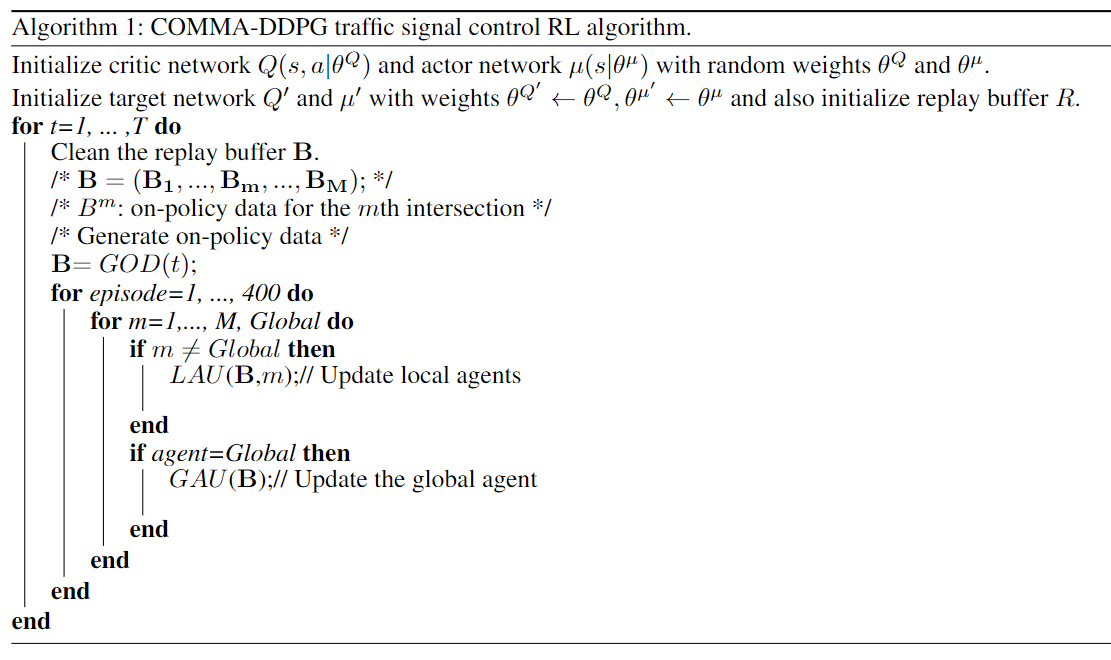}
% \caption{Map of 160 intersections}
% \label{fig:alg1}
\end{figure}

\begin{figure}[H]
\centering
\includegraphics[scale = 0.55]{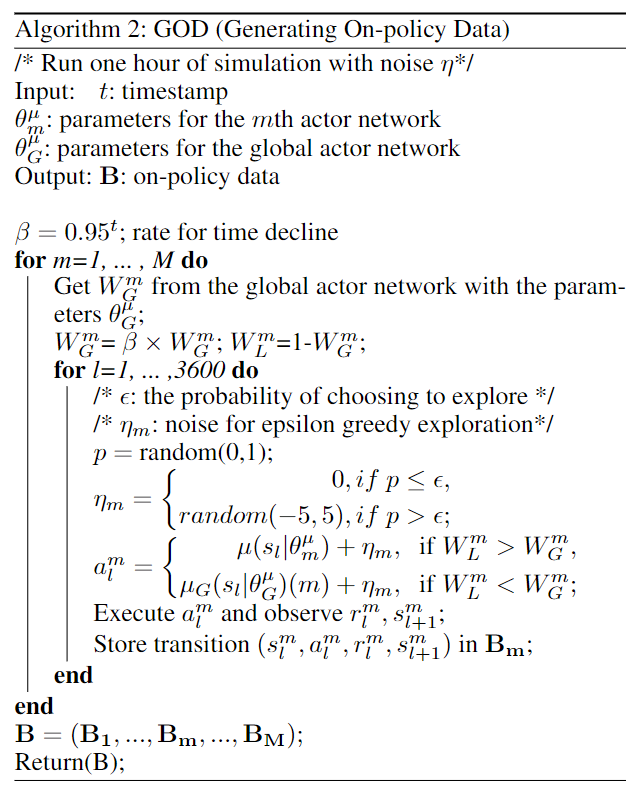}
% \caption{Map of 160 intersections}
% \label{fig:alg1}
\end{figure}

\begin{figure}[H]
\centering
\includegraphics[scale = 0.55]{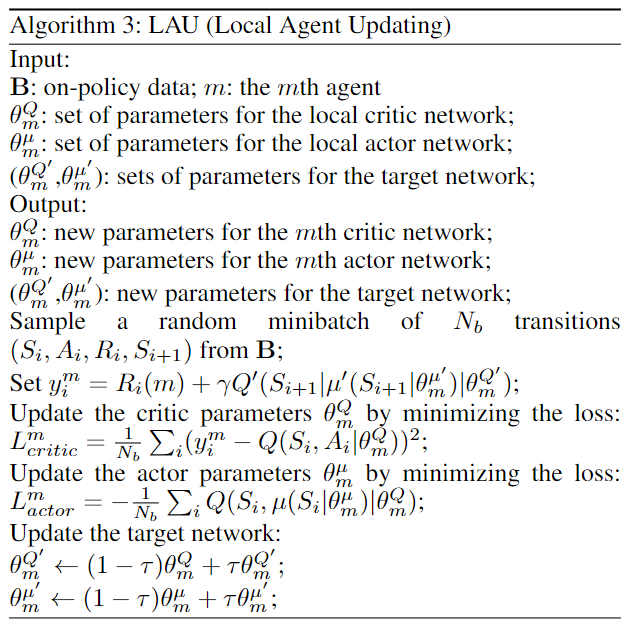}
% \caption{Map of 160 intersections}
% \label{fig:alg1}
\end{figure}

\begin{figure}[H]
\centering
\includegraphics[scale = 0.55]{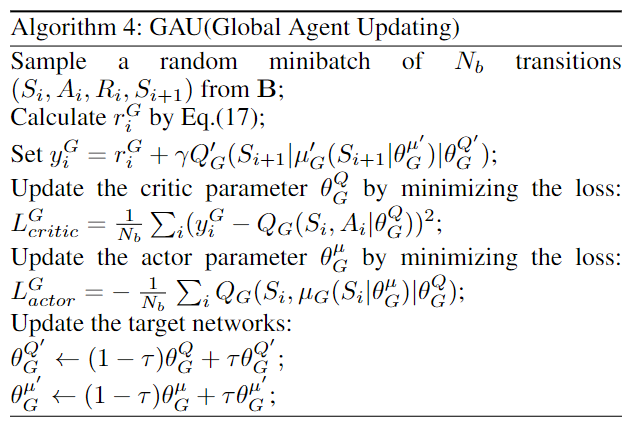}
% \caption{Map of 160 intersections}
% \label{fig:alg1}
\end{figure}
\subsection{Experimental Table}
\begin{table}[h] %table 5
% \tiny
\footnotesize
\setlength\tabcolsep{3pt}
\centering
\caption{ Performance comparisons among different SoTA methods when ten intersections were included.}
\vspace{0.2cm}
\begin{tabular}{lcccccc}%lcc代表三欄，第一欄靠左，二三欄置中，兩旁的直線代表欄與欄間劃上直線
\hline  %劃上一條橫線
    Methods & Delay & Speed & Time loss & Travel time & Waiting time\\\hline
    IDQN [Ault and Sharon, 2021] & 1814.18 & 10.54 & 771.29 & 517.82 & 669.59\\
    IPPO [Ault and Sharon, 2020]  & 1861.14 & 9.71 & 1371.08 & 664.55 & 1201.17\\
    FMA2C [Chu et al., 2016]  & 1784.95 & 10.61 & 668.03 & 512.62 & 569.59\\
    MPLight [Zheng et al., 2019] & 1750.39 & 10.69 & 569.07 & 523.38 & 492.83\\
    MPLight full(MPLight+IDQN) & 1685.54 & 10.91 & 431.79 & 520.78 & 324.46\\\hline
    COMMA-DDPG & 603.94 & 11.06 & 202.03 & 493.71 & 204.59 \\\hline
\end{tabular}%\\\\

\label{tab:ComarionsBenchmark10Intersections}
% \vspace{-0.5cm}
\end{table}

\begin{table}[h] %table 5
% \tiny
\footnotesize
\setlength\tabcolsep{3pt}
\centering
\caption{Performance of our method on 160 intersections.}
\vspace{0.2cm}
\begin{tabular}{lcccccc}%lcc代表三欄，第一欄靠左，二三欄置中，兩旁的直線代表欄與欄間劃上直線
\hline  %劃上一條橫線
    Methods &  avg. Waiting time & Fuel(mg/s) & CO(mg/s) & CO2(mg/s) \\\hline
    Our methods & 494.53 & 1.01 & 103.62 & 2335.86  \\\hline
    No global agent & 528.77 & 1.06 & 107.3 & 2356.81 \\\hline
\end{tabular}%\\\\

\label{tab:Performance on 160 intersections}
% \vspace{-0.5cm}
\end{table}
% \newpage
\begin{table}[h]
% \footnotesize
\setlength\tabcolsep{3pt}
\centering
\caption{Measured unit of variables.}
\vspace{0.2cm}
\begin{tabular}{|c|c|}
\hline 
Variables & Measured Unit\\
\hline
$CO_{engine}$ & g/kWh \\
\hline
$V_{engine}$ & L \\
\hline
FC & L/100km \\
\hline
$M_{fuel}$ & g/mole \\
\hline
$M_{air}$ & g/mole \\
\hline
$r_{stop}$ & kW \\
\hline
$t_{stop}$ & s \\
\hline 
\end{tabular} \\
\label{tab:my_label}
\end{table}
Table~\ref{tab:ComarionsBenchmark10Intersections} presents performance comparisons using ten intersections to construct the road networks. Among all state-of-the-art (SoTA) methods, IPPO exhibits significant instability and performs poorly. When evaluating a subset of intersections, FMA2C outperforms other methods. However, as the number of intersections increases, MPLight-related methods demonstrate their superiority in traffic signal control. For instance, although the "MPLight-full" method initially performs worse than other methods with a smaller number of intersections, it surpasses most SoTA methods in various performance categories. Except for our proposed method, it achieves the best performance across categories such as "Delay," "Speed," "Time loss," and "Waiting time." Notably, leveraging a global agent, our COMMA-DDPG method outperforms all SoTA methods in all performance categories.\\

Table~\ref{tab:Performance on 160 intersections} presents a comparison method after processing the global agent in parallel. The table consists of a 5-by-5 global agent data for 169 intersections(as shown in Figure \ref{fig:160intersection}). In our simulation, some intersections are designed as T-shaped or I-shaped intersections, representing real-world scenarios where only left and right or up and down movements are allowed. Specifically, there are 9 I-shaped intersections, and their corresponding left and right (or upper and lower) intersections become T-junctions. Due to the unique shape of these intersections, some positions in the 5 by 5 global agent grid are missing. To handle this, we simply assign a value of zero to these vacant positions. \\

Table~\ref{tab:my_label} provides the unit representation of certain parameters used in the formula to calculate carbon emissions referenced in this article. The variable CO can be replaced with CO2 to calculate CO2 emissions. The formula consists of the following parameters:
\begin{itemize}
\item  $CO_{engine}$: CO emission from the vehicle engine in the driving state.
\item $V_{engine}$: Exhaust volume of the engine.
\item FC: Fuel consumption of the vehicle.
\item $M_{fuel}$: Molecular weight of the fuel.
\item $M_{air}$: Molecular weight of the air.
\item $r_{stop}$: Average power of the vehicle when it is stopped.
\item $t_{stop}$: Duration of time when the vehicle is stopped.
\end{itemize}
It should be noted that, except for $t_{\text{stop}}$, which is influenced by our experiment, the remaining parameters are not affected as long as the same traffic flow is used for simulation.

%%%%%%%%%%%%%%%%%%%%%%%%%%%%%%%%%%%%%%%%%%%%%%%%%%%%%%%%%%%%%%%%%%%%%%%%%%%%%%%
% \newpage
\subsection{Picture}
% \newpage
% \begin{figure}[h]
% \includegraphics[scale=0.4]{picture1 (1).pdf}
% \end{figure}

\begin{figure}[h]
% \vspace{3cm}
     \centering
     \includegraphics[scale=0.22]{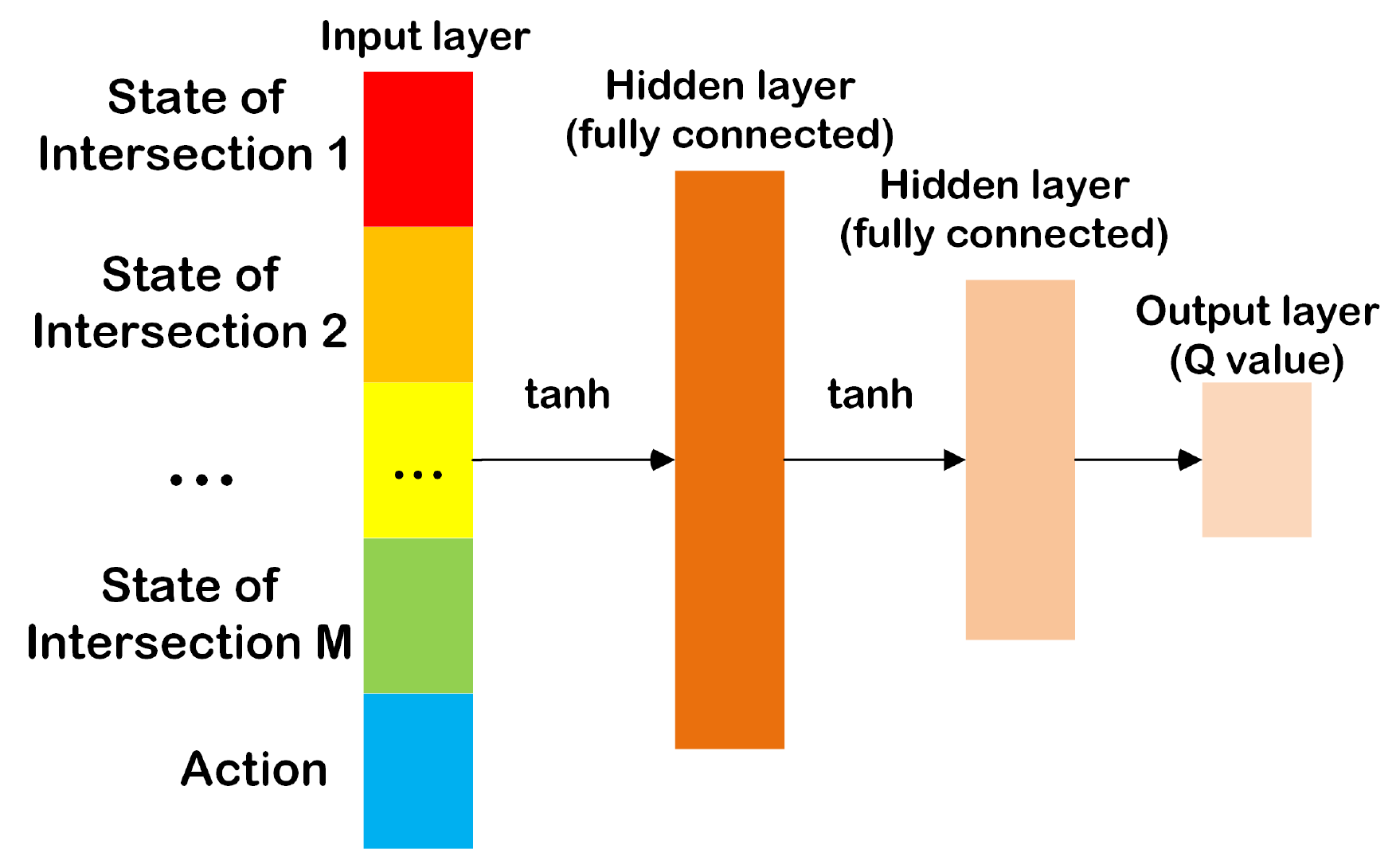}
     \centering
     \includegraphics[scale=0.22]{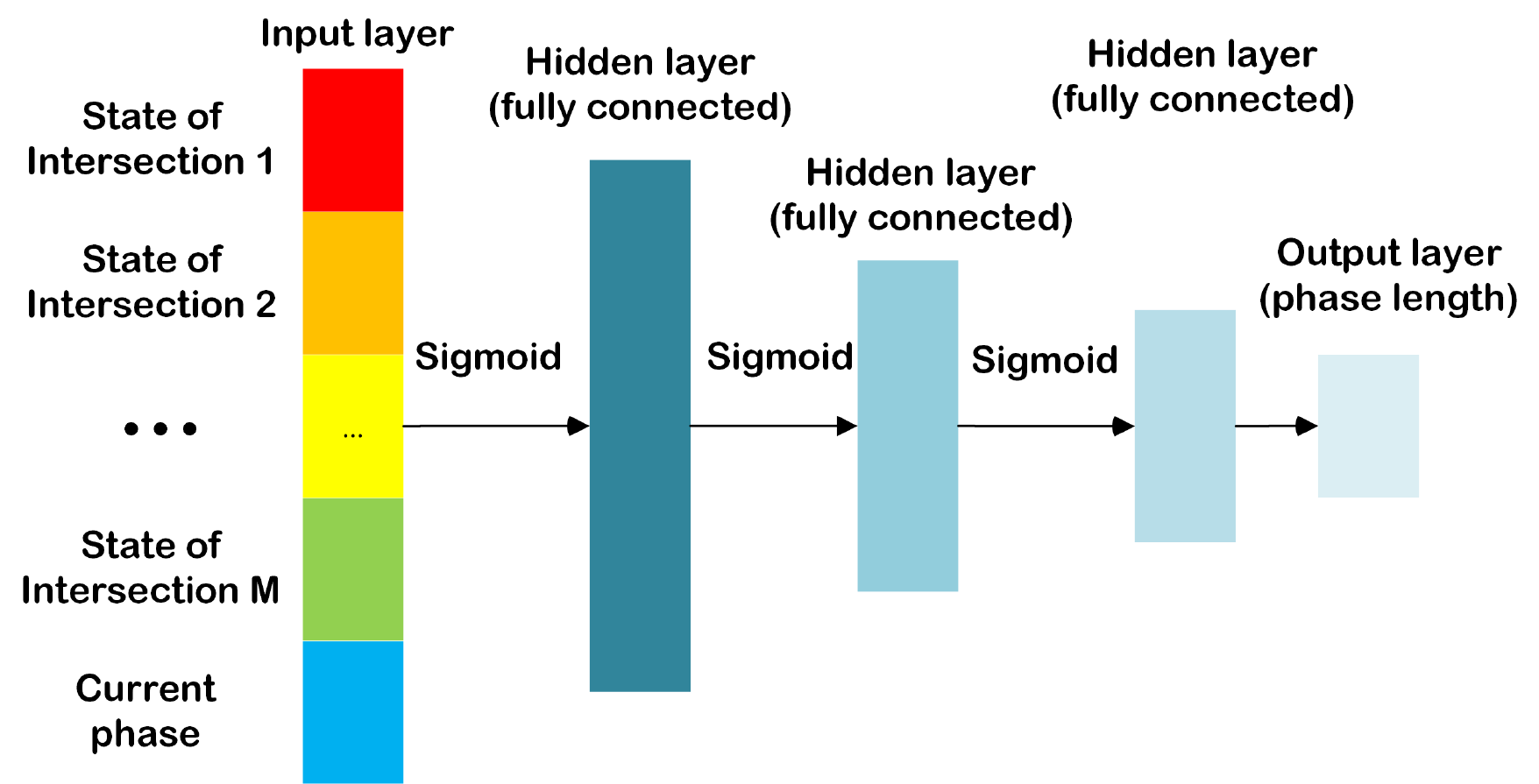}
      \qquad  \qquad  \qquad \footnotesize (a) \qquad \qquad \qquad \qquad \qquad \qquad \footnotesize (b)
     \caption{Architectures for local agent. (a) Local critic.(b) Local actor.}
     \label{fig:local_agent}
 
 \end{figure}

% \vspace{-2cm}
 \begin{figure}[h]
 % \centering
 
 \includegraphics[scale=0.22]{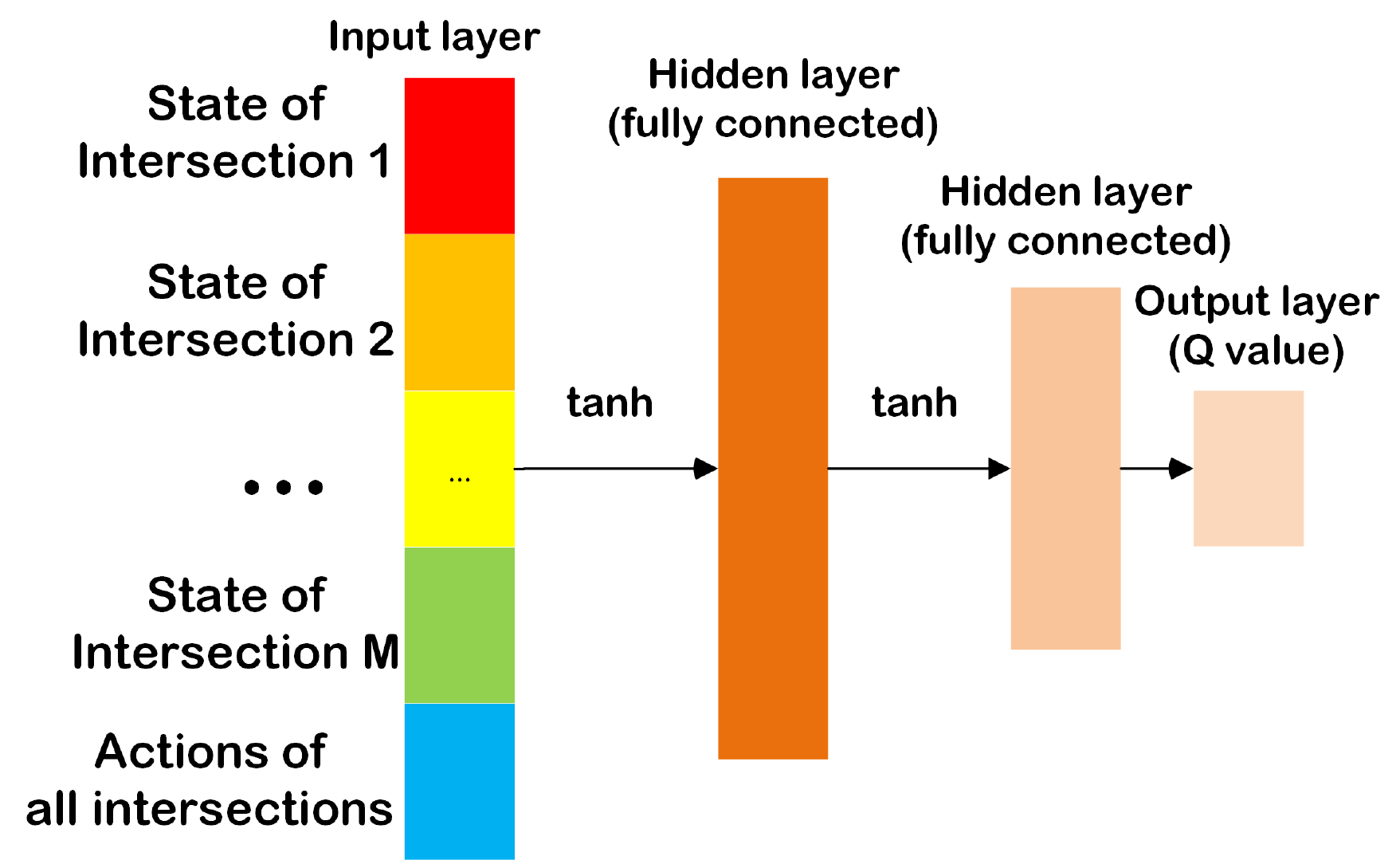}
 \centering
 \includegraphics[scale=0.22]{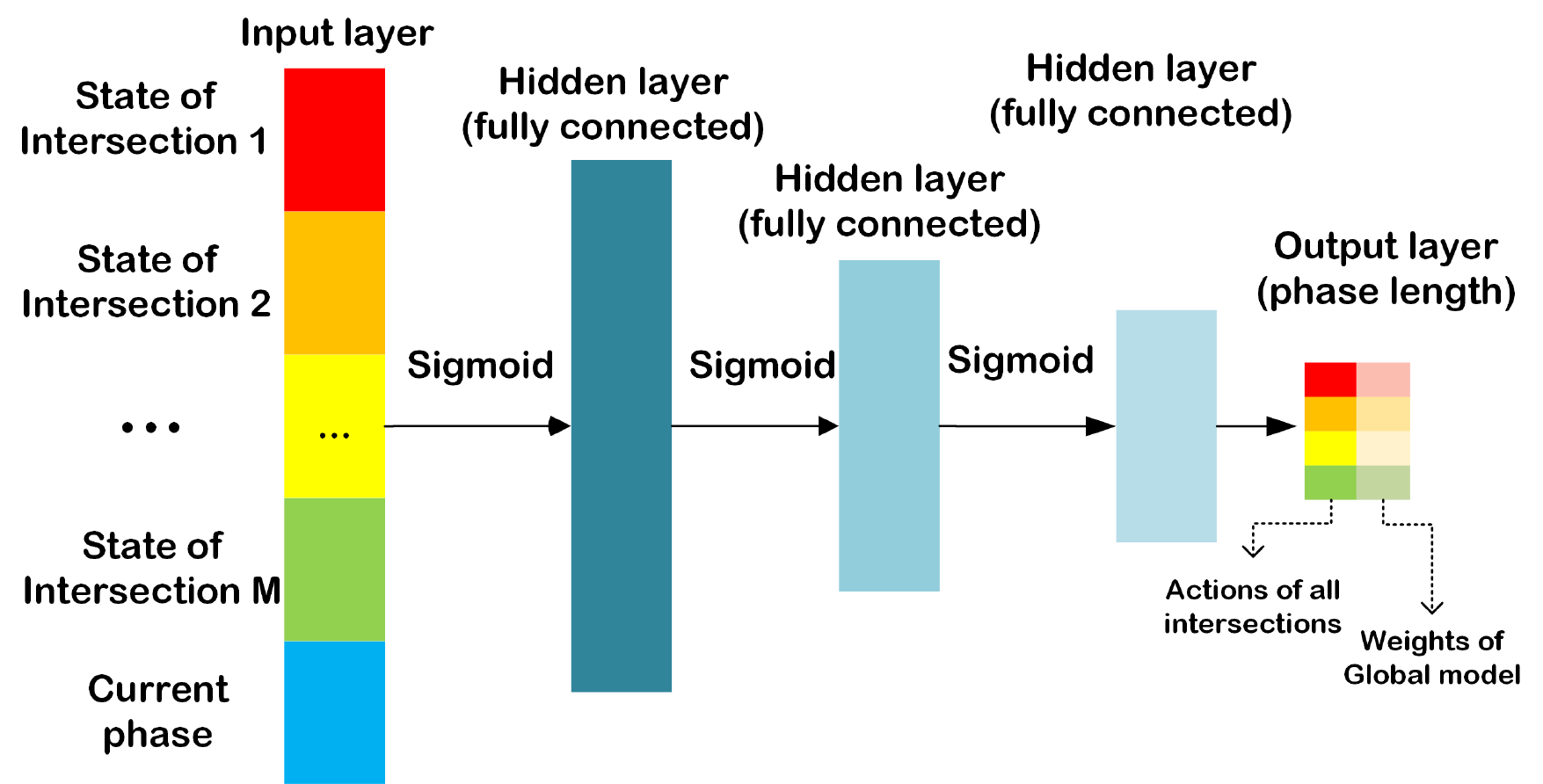}
   \qquad  \qquad  \qquad \footnotesize (a) \qquad \qquad \qquad \qquad \qquad \qquad \footnotesize (b)
 \caption{Architectures for global agent. (a) Global critic.(b) Global actor.}
 \label{fig:global_agent}
\end{figure}
% \newpage

% \vspace{-1000000cm}
\begin{figure}[h]
% \centering
    \centerline{
    %  (a) \includegraphics[scale=0.45]{LaTeX/road.png}
     (a) \includegraphics[scale=0.17]{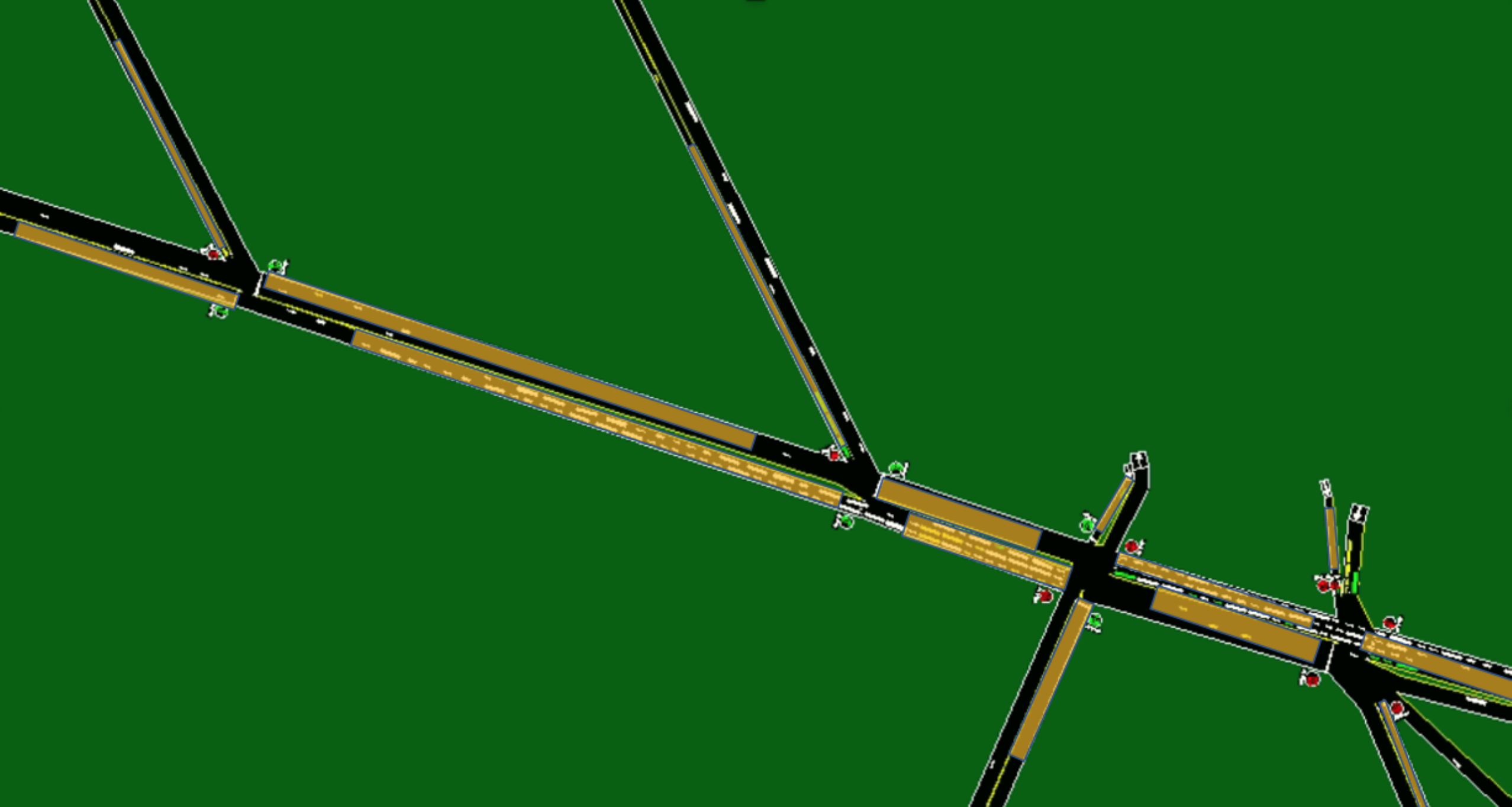}
    } 
    \centerline{
     (b) \includegraphics [scale=0.17]{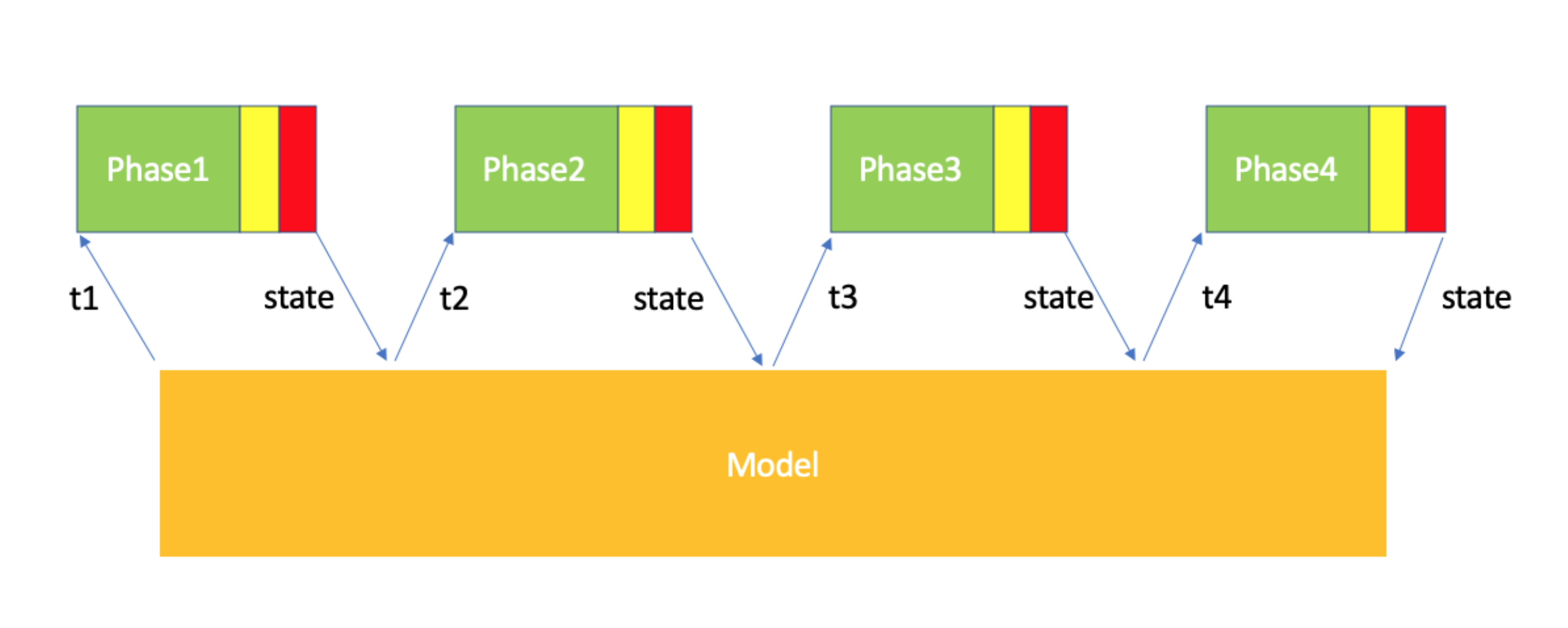}
    }
    \caption{
    (a) Simulation traffic environment for RL traffic control, where yellow area is the visible range of each lane. 
    (b) Control process. Here, $t_{i}$ means the duration of green light of phase $i$.
    }
    \label{fig:overview}
\end{figure}
% \vspace{-10cm}
\begin{figure}[h]
\centering
\includegraphics[scale = 0.4]{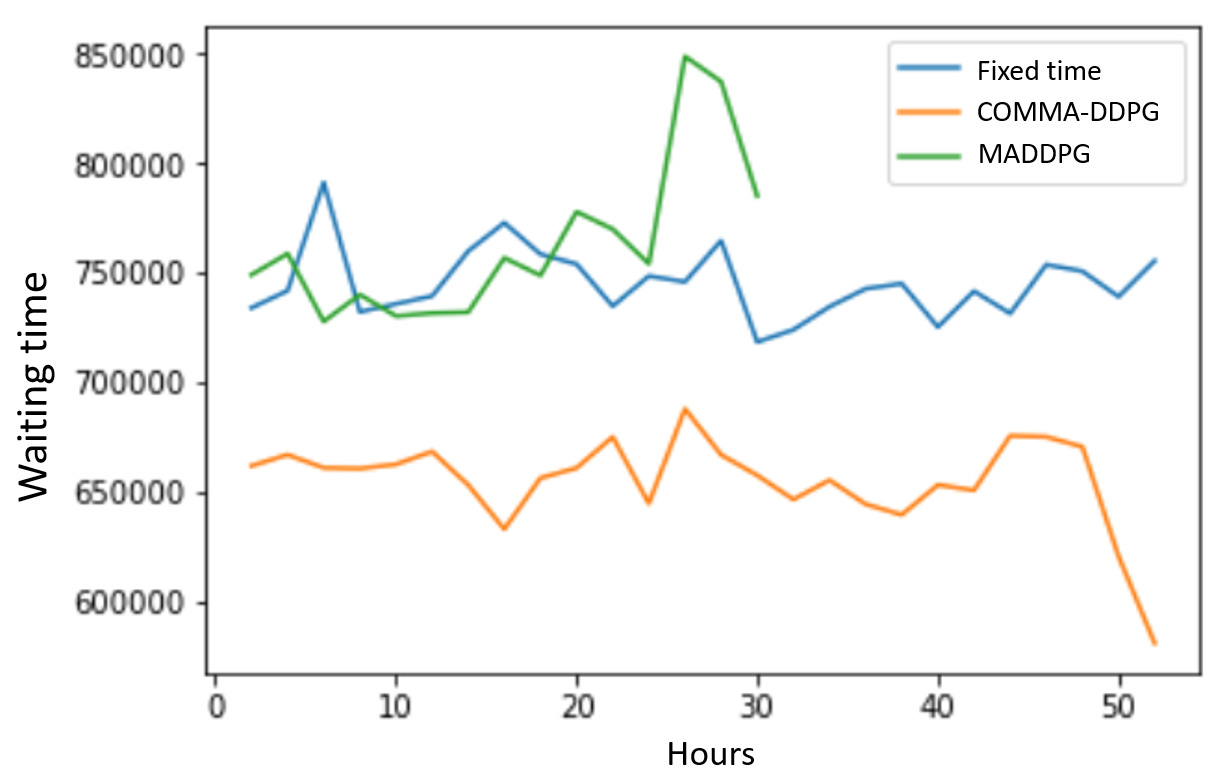}
\caption{Waiting time converge conditions during the training process among different methods.}
\label{fig:WaitingTime}
\end{figure}

\begin{figure}[h]
\centering
\includegraphics[scale = 0.4]{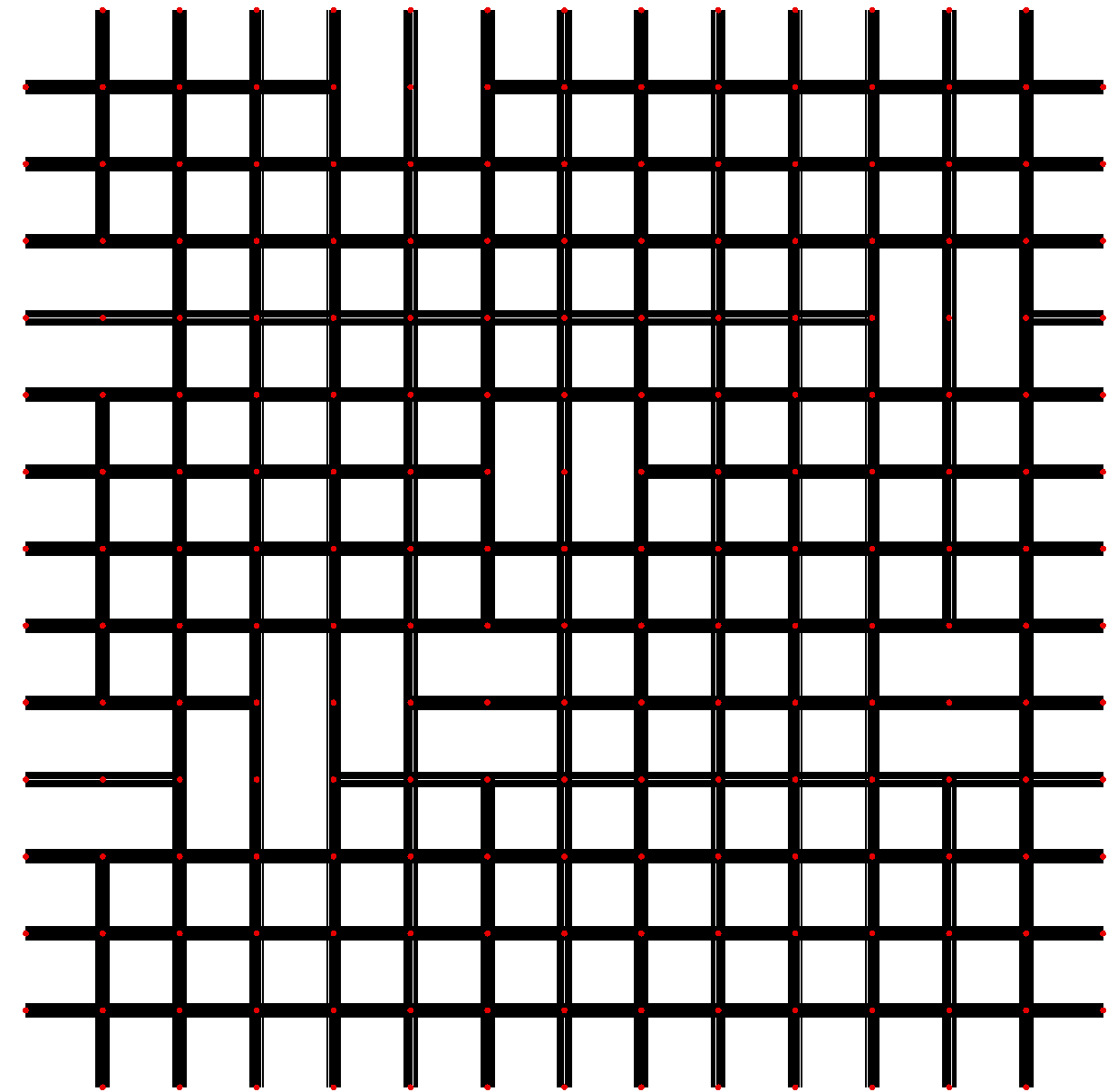}
\caption{Map of 160 intersections}
\label{fig:160intersection}
\end{figure}

\end{document}